\documentclass[nonacm]{acmart}
\usepackage{graphicx}
\usepackage{latexsym}
\usepackage{amsmath}
\usepackage{booktabs}
\usepackage{enumitem}

\setcopyright{acmcopyright}
\copyrightyear{2023}
\acmYear{2023}
\acmDOI{XXXXXXX.XXXXXXX}

\acmJournal{TKDD}
\acmVolume{1}
\acmNumber{1}
\acmArticle{1}
\acmMonth{1}

\begin{document}

\title{Properties of fairness measures in the context of varying class imbalance and protected group ratios}

\author{Dariusz Brzezinski}
\author{Julia Stachowiak}
\author{Jerzy Stefanowski}
\author{Izabela Szczech}
\author{Robert Susmaga}
\author{Sofya Aksenyuk}
\author{Uladzimir Ivashka}
\author{Oleksandr Yasinskyi}
\affiliation{%
  \institution{Institute of Computing Science, Poznan University of Technology}
  \streetaddress{ul. Piotrowo 2}
  \postcode{60-965}
  \city{Poznan}
  \country{Poland}
}
\email{dariusz.brzezinski@cs.put.poznan.pl}
\renewcommand{\shortauthors}{Brzezinski et al.}

\begin{abstract}
Society is increasingly relying on predictive models in fields like criminal justice, credit risk management, or hiring. To prevent such automated systems from discriminating against people belonging to certain groups, fairness measures have become a crucial component in socially relevant applications of machine learning. However, existing fairness measures have been designed to assess the bias between predictions for protected groups without considering the imbalance in the classes of the target variable. Current research on the potential effect of class imbalance on fairness focuses on practical applications rather than dataset-independent measure properties. In this paper, we study the general properties of fairness measures for changing class and protected group proportions. For this purpose, we analyze the probability mass functions of six of the most popular group fairness measures. We also measure how the probability of achieving perfect fairness changes for varying class imbalance ratios. Moreover, we relate the dataset-independent properties of fairness measures described in this paper to classifier fairness in real-life tasks. Our results show that measures such as Equal Opportunity and Positive Predictive Parity are more sensitive to changes in class imbalance than Accuracy Equality. These findings can help guide researchers and practitioners in choosing the most appropriate fairness measures for their classification problems.
\end{abstract}



\maketitle

\section{Introduction}
Machine learning systems are increasingly being used to make decisions that affect people. Despite the benefits often associated with improving or speeding up tasks, there is also an awareness of the risks associated with such systems. In particular, there is a general agreement that machine learning models need to be controlled to maintain fairness and avoid biased decisions. These issues are reflected in recent AI guidelines, such as the proposed EU regulation on AI~\cite{euaiact} and the recently accepted UNESCO recommendation on ethics in AI~\cite{unesco}. The fairness of machine learning models is also a key research driver toward Responsible and Trustworthy AI~\cite{trustworthyai}.

\textit{Fairness} in machine learning refers to the idea that predictive systems should be designed and operated in a way that is fair and just to all individuals and groups~\cite{acm_review}. This means that machine learning models should not discriminate against people based on their race, gender, age, or other personal characteristics. Therefore, approaches that mitigate unfairness are based on the notion of \textit{protected attributes} (sometimes also called sensitive attributes) that define \textit{protected and unprotected groups}, i.e., groups that are disproportionately less or more likely to be positively classified. Practical machine learning applications with protected attributes include automated hiring procedures~\cite{hiring}, credit scoring~\cite{creditrisk}, and criminal justice~\cite{compas}.

In this context, several measures that assess fairness towards protected groups have been put forward~\cite{acm_review}. However, current research focuses mainly on proposing methods that improve the fairness of machine learning models~\cite{friedler2019comparative} or studying correlations between fairness measures in practical applications~\cite{anahideh2021choice}, rather than investigating general dataset-independent properties. Therefore, providing general advice on which fairness measures are best suited for a given case study is hard. Moreover, by focusing on dataset-oriented evaluations, current studies do not answer questions concerning the behavior of fairness measures in the presence of different types of bias in the data (representation bias, skewed distributions, feature bias). In particular, there are no studies connecting theoretical properties of fairness measures with different levels of class imbalance, i.e., situations where at least one of the target classes contains a much smaller number of examples than the other classes~\cite{he2009learning}. Such a study would be valuable in guiding machine learning practitioners on the use of fairness measures, as many real-world datasets are naturally imbalanced~\cite{branco2016survey}. Finally, there has been no investigation into how interactions between class imbalance and disproportions between protected and unprotected groups might affect the properties of fairness measures.

In this paper, we define general dataset-independent properties of fairness measures that assess their behavior for varying levels of class imbalance and protected group bias. To identify these properties in popular group fairness measures, we analyze their distributions in the context of class and protected group imbalance. As a result, we provide guidelines on which fairness measures are applicable to which types of datasets. The detailed contributions of this paper are as follows:
\begin{itemize}
    \item In Sections~\ref{sec:measures} and \ref{sec:distributions}, we recall six popular group fairness measures and put forward a \textbf{method for analyzing fairness measures} on the basis of their probability mass functions.
    \item In Section~\ref{sec:properties}, we propose \textbf{a set of general (dataset-independent) fairness measure properties} related to their behavior in the presence of different levels of class imbalance and protected group ratios. We then verify whether the studied six fairness measures possess the proposed properties.
    \item In Section~\ref{sec:case-study}, by training six different classifiers in a controlled experiment \textbf{case study using real-world data with varying class and protected group ratios}, we verify whether the identified dataset-independent properties apply to practical classification scenarios.
    \item In Section~\ref{sec:conclusions}, we formulate \textbf{guidelines on using fairness measures} in different data scenarios and draw lines for future research.
\end{itemize}

\section{Related work}
\label{sec:related-work}

The concepts discussed in this paper, fall into the field of machine learning fairness ~\cite{caton2020fairness,Dunkelau2020,Romei2013AMS}. Although the general notion of fairness has been thoroughly studied for several decades, e.g., in social sciences, the awareness of fairness in the machine learning community has developed only in the last few years. Since then, there has been growing interest in the topic, which has led to proposing different fairness notations and constructing many new approaches to prevent classifiers from discriminating against certain groups of people~\cite{zhou2021bias}.

It is worth noting that there is no single definition of fairness and no single way of verifying it mathematically. For instance, Dablain \textit{et al.} mention as many as 21 notions of fairness~\cite{dablain2022towards}. Moreover, these definitions of fairness can be categorized in different ways~ \cite{castelnovo2022clarification,Dunkelau2020,gajane2018formalizing,friedler2019comparative}. For example, according to the categorization by Gajane and Pechenizkiy~\cite{gajane2018formalizing}, one can formalize group fairness (considered in this paper), individual fairness, preference of treatment, preference of impact, equality of opportunity, counterfactuals, and unawareness. Among all these options, the literature does not determine which approach to fairness is the best. The two most popular categories are group fairness and individual fairness measures. Group fairness can be quantified by means of differences or ratios. In this study, we follow Žliobaitė~\cite{Zliobait2017MeasuringDI}, who recommends using difference-based rather than ratio-based definitions. Finally, several papers discuss the orthogonality of various fairness measures~\cite{castelnovo2022clarification}. It has been shown that it is impossible to satisfy all fairness measures simultaneously since many of them are mutually exclusive~\cite{corbettdavies2018measure}.

Most research efforts in machine learning fairness have been directed at proposing methods for detecting~\cite{baniecki2021dalex} and correcting~\cite{caton2020fairness} discriminatory bias of machine learning models. Such fairness interventions to models are usually divided into three categories: (1) \textit{pre-processing} (performing specific transformations of data to remove discrimination bias from the training data); (2) \textit{in-processing} (modifying the model's training to adhere to some fairness criteria or constraints); (3) \textit{post-processing} (correcting the biased classifier with respect to the protected attribute or selected measures). For more details on classifier intervention methods, the reader is referred to surveys of Caton and Haas~\cite{caton2020fairness} and Dunkelau~\cite{Dunkelau2020}. 

Studies on machine learning fairness usually do not explicitly take into account the relative sizes and imbalances of classes and protected groups despite the fact that such characteristics were often present in the data under consideration. On the other hand, it has been observed that the presence of class imbalance may not only lead to poor recognition of minority classes but also to the worsening of fairness measures \cite{dablain2022towards}. Also, Deng \textit{et al.}~\cite{deng2022fifa} have reported experiments with imbalanced datasets confirming the potential of class imbalance impacting the fairness of classifier predictions.

Works that tie fairness with the large body of work on class imbalance~\cite{fernandez2018learning} mainly introduce new specialized methods for improving the selected fairness measure while simultaneously dealing with class imbalances when training the classification models. Kamiran and Calders~\cite{Kamiran2009} proposed a pre-processing method, which makes the least intrusive modifications to the proportions of protected and unprotected examples that lead to an unbiased dataset and
non-discriminating classifier. Iosifidis and Ntoutsi~\cite{Iosifidis2020FABBOOO} studied online learning of classifiers in the presence of both biases and proposed to modify boosting ensembles. Their proposal involves adjusting the weights of new training examples according to the monitored changes of the imbalance ratio and adaption of the final decision threshold with respect to one of two fairness measures. Another approach is presented by Ferrari and Bacciu~\cite{ferrari2021addressing}, where the authors introduce a new loss function that takes into account both fairness and class imbalance. Similarly, a logit-based loss function that incorporates class imbalance-dependent margins was studied in~\cite{deng2022fifa}. In parallel, fair empirical risk minimization provides a framework for optimizing classifiers to conform with a given fairness-oriented loss function~\cite{fairrisk1, fairrisk2}. 
Finally, a resampling-based pre-processing approach, named FairOversampling, has been recently proposed in~\cite{dablain2022towards}.

Nevertheless, to the best of our knowledge, there is still a lack of systematic research on the impact of different levels of class imbalance and protected group ratios on the properties of selected fairness measures. Aiming to fill this gap, our study builds on:
\begin{itemize}
    \item[(A)] works defining fairness measures for different applications \cite{castelnovo2022clarification,caton2020fairness,Dunkelau2020,Zliobait2017MeasuringDI}, 
    \item[(B)] studies on class imbalance and fairness \cite{dablain2022towards,deng2022fifa}, 
    \item[(C)] our previous works on analyzing classification measures using confusion matrices~\cite{DBLP:journals/isci/BrzezinskiSSS18,brzezinski2019dynamics}. 
\end{itemize}
In this paper, we investigate fairness measures defined in (A) in the context of problems discussed in (B) using methods inspired by (C). The novelty of our approach comes from using probability mass functions (C) in a new setting (A+B) for pairs of confusion matrices within subgroups defined by imbalance and group ratios. We deviate from existing analyses on group fairness measures in (B) because we are independent of concrete datasets and classifiers and because we focus on general properties. We believe that the analysis of confusion matrices provides a broader view compared to results based on individual datasets.

Our work asks the question of \textit{which} fairness measure should be optimized and \textit{when}, whereas most existing works focus on \textit{how} to enforce selected measures. Therefore, our work complements existing research on fairness intervention methods, showing the (statistical) suitability of measures for particular types of datasets and classification scenarios. That being said, it is important to acknowledge that fairness is a sensitive topic. Therefore, in real-world applications, several additional aspects (e.g., legal, social, or cultural) and biases should be very carefully considered when choosing appropriate metrics. Critical discussions of the differences between the technical approach to machine learning fairness and the legal aspects of regulations can be found in recent works by Wachter et al.~\cite{wachter2020bias} and Kirat et al.~\cite{kirat2022fairness}.

\section{Group fairness measures}
\label{sec:measures}

To tackle the problem of potentially discriminatory behavior of machine learning models, researchers have put forward many ways of quantifying fairness~\cite{caton2020fairness,castelnovo2022clarification,Dunkelau2020}. Among many different notions of fairness, the most commonly used metrics emphasize either \textit{individual} (similarity-based) fairness or \textit{group} (statistical) fairness. In this paper, we will focus on group fairness, which underlines equal treatment of various groups identified by protected attributes~\cite{castelnovo2022clarification}. Without loss of generality, for the sake of simplicity, we will consider problems with one protected categorical attribute that divides examples in a classification dataset into two groups: the \textit{protected group} ($_{p}$) and the \textit{unprotected group} ($_{\mathit{up}}$). Moreover, we will focus on fairness measures that can be defined for binary classification problems, where the two considered classes will be referred to as the \textit{postive class} and the \textit{negative class}. The numbers of positive and negative examples in a dataset will be denoted as $P$ and $N$, respectively. The total number of examples in a dataset will be denoted as $n = P + N$.

\begin{figure}[htb]
\centerline{
\includegraphics[width=\textwidth]{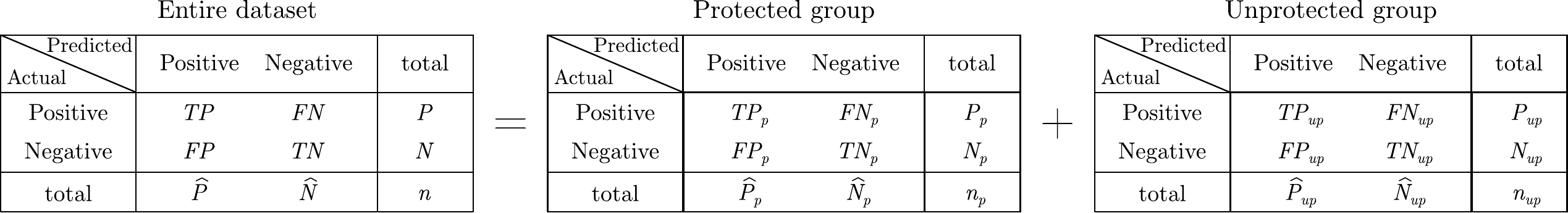}
}
\caption{Confusion matrix for two-class classification. The diagram shows how, in the context of quantifying fairness, a confusion matrix can be viewed as the sum of two confusion matrices---one for the protected group and one for the unprotected group.} \label{fig:confusion-matrix}
\end{figure}

Following the above, group fairness measures can be defined using entries from a two-class confusion matrix presented in Figure~\ref{fig:confusion-matrix}. The $\mathit{TP}$ (\textit{True Positive}) and $\mathit{TN}$ (\textit{True Negative}) entries denote the number of examples classified correctly by the classifier as positive and negative, whereas the $\mathit{FN}$ (\textit{False Negative}) and $\mathit{FP}$ (\textit{False Positive}) indicate the number of misclassified positive and negative examples, respectively~\cite{japkowicz2011evaluating}. In the context of group fairness, this definition of a confusion matrix can be considered as a sum of confusion matrices for the protected and unprotected group, with entries indexed with $_{p}$ and $_{\mathit{up}}$, respectively (Figure~\ref{fig:confusion-matrix}). For the purpose of quantifying fairness, when referring to a confusion matrix, we will imply a tuple of eight values: $\{\mathit{TP}_{p}, \mathit{FN}_{p}, \mathit{FP}_{p}, \mathit{TN}_{p}, \mathit{TP}_{up}, \mathit{FN}_{up}, \mathit{FP}_{up}, \mathit{TN}_{up}\}$.

A dataset can have different proportions of examples belonging to positive/negative classes and to protected/unprotected groups. To quantify the class imbalance and group imbalance in a given dataset consisting of $n$ examples, we will use the notions of \textit{imbalance ratio} ($IR$) and \textit{group ratio} ($GR$) defined as:
\begin{align*}
\mathit{IR} &= \frac{P}{n} & \mathit{GR} &= \frac{n_p}{n}
\end{align*}
Using the above notation, a dataset with perfect class balance will have $\mathit{IR}=0.5$. Independently, a dataset with an even number of examples in the protected and unprotected groups will have $\mathit{GR}=0.5$.

In our work, we focused on fairness measures that can be expressed by the eight entries in the confusion matrices defined above. Barocas et al. denote such measures as \textit{observational}~\cite{barocas2023fairness}. More formally, we will analyze measures that depend on:
\begin{itemize}
    \item the protected (sensitive) attribute $S$ that defines the groups for which we want to measure fairness ($g_p$, $g_{up}$);
    \item the target attribute $Y$, which in binary classification represents two classes that we can predict ($Y = 0$ or $Y = 1$);
    \item the classification score $C$, which represents the predicted score within [0, 1]. 
\end{itemize}

Within observational group fairness measures, we selected six that are traditionally defined as equalities and cover all three categories of non-discrimination criteria: independence ($C  \, \bot \,  S$), separation ($C  \, \bot \,  S|Y$), sufficiency ($Y \, \bot \, S|C$)~\cite{caton2020fairness,barocas2023fairness}. 
From the \textit{independence} category, we chose \textit{Accuracy Equality} (Eq.~\ref{eq:ae_prob})~\cite{compas} and \textit{Statistical Parity}  (Eq.~\ref{eq:sp_prob})~\cite{zemel2013learning}.
\begin{equation}
    P(C=Y|S=g_p) = P(C=Y|S=g_{up})\label{eq:ae_prob}
\end{equation}
\begin{equation}
    P(C=1|S=g_p) = P(C=1|S=g_{up})\label{eq:sp_prob}
\end{equation}
From the \textit{separation} category, we chose \textit{Equal Opportunity} (Eq.~\ref{eq:eo_prob})~\cite{hardt2016equality} and \textit{Predictive Equality} (Eq.~\ref{eq:pe_prob})~\cite{Corbett-Davies_2017}.
\begin{equation}
    P(C=1|Y=1,S=g_p) = P(C=1|Y=1,S=g_{up})\label{eq:eo_prob}
\end{equation}
\begin{equation}
    P(C=1|Y=0,S=g_p) = P(C=1|Y=0,S=g_{up})\label{eq:pe_prob}
\end{equation}
Finally, from the \textit{sufficiency} category we chose \textit{Positive Predictive Parity} (Eq.~\ref{eq:ppp_prob}) and \textit{Negative Predictive Parity} (Eq.~\ref{eq:npp_prob})~\cite{chouldechova2017fair}.
\begin{equation}
    P(Y=1|C=1,S=g_p) = P(Y=1|C=1,S=g_{up})\label{eq:ppp_prob}
\end{equation}
\begin{equation}
    P(Y=0|C=0,S=g_p) = P(Y=0|Y=0,S=g_{up})\label{eq:npp_prob}
\end{equation}

Even though the notions of group fairness are traditionally discussed in terms of probabilities, for concrete classification scenarios these probabilities are usually estimated using the entries from the confusion matrix. Moreover, in order to quantify fairness, these probability equations are transformed into either ratios or differences. In this study, we have chosen the latter form of quantifying fairness as the values of differences are constrained to a $[-1, 1]$ range, less prone to division by zero problems, and easier to interpret~\cite{Zliobait2017MeasuringDI}. Moreover, definitions based on differences are more general than definitions based on equations. In fact, the satisfaction of a fairness equation corresponds to a difference being equal to zero; we will refer to this special case as \textit{perfect fairness}. 

By redefining the selected measures as differences based on entries of confusion matrices we get:
\begin{align}
\text{Accuracy Equality} &= \frac{\mathit{TP}_{p} + \mathit{TN}_{p}}{n_{p}} - \frac{\mathit{TP}_{\mathit{up}} + \mathit{TN}_{\mathit{up}}}{n_{\mathit{up}}}\label{eq:ae}\\
\text{{Statistical Parity}} &= \frac{\mathit{TP}_{p} + \mathit{FP}_{p}}{n_{p}} - \frac{\mathit{TP}_{\mathit{up}} + \mathit{FP}_{\mathit{up}}}{n_{\mathit{up}}}\label{eq:sp}\\
\text{{Equal Opportunity}} &= \frac{\mathit{TP}_{p}}{\mathit{FN}_{p} + \mathit{TP}_{p}} - \frac{\mathit{TP}_{\mathit{up}}}{\mathit{FN}_{\mathit{up}} + \mathit{TP}_{\mathit{up}}}\label{eq:eo}
\end{align}
\begin{align}
\text{{Predictive Equality}} &= \frac{\mathit{FP}_{p}}{\mathit{FP}_{p} + \mathit{TN}_{p}} - \frac{\mathit{FP}_{\mathit{up}}}{\mathit{FP}_{\mathit{up}} + \mathit{TN}_{\mathit{up}}}\label{eq:pe}\\
\text{{Positive Predictive Parity}} &= \frac{\mathit{TP}_{p}}{\mathit{FP}_{p} + \mathit{TP}_{p}} - \frac{\mathit{TP}_{\mathit{up}}}{\mathit{FP}_{\mathit{up}} + \mathit{TP}_{\mathit{up}}}\label{eq:ppp}\\
\text{{Negative Predictive Parity}} &= \frac{\mathit{TN}_{p}}{\mathit{FN}_{p} + \mathit{TN}_{p}} - \frac{\mathit{TN}_{\mathit{up}}}{\mathit{FN}_{\mathit{up}} + \mathit{TN}_{\mathit{up}}}\label{eq:npp}
\end{align}

\section{Distribution-based analyses of fairness measures}
\label{sec:distributions}

In this study, our goal is to examine the behavior of fairness measures in the context of varying class imbalance and protected group ratios. As shown in the previous section, the values of the analyzed fairness measures are derived from confusion matrices, which represent the results of classification on experimental data. By considering the training data as an outcome of a random process, we can provide a probabilistic view of fairness measure values. Specifically, measures based on confusion matrices can be regarded as discrete random variables that map a confusion matrix into a numerical value. Discrete random variables are typically characterized by their \textit{probability mass functions} ($\mathit{pmfs}$), which denote the probability that a discrete random variable precisely equals a specific value~\cite{soong04}. Probability mass functions are frequently represented as histograms, with the $x$-axis indicating measure values and the $y$-axis signifying the probability of obtaining a particular value. We will employ such visualizations to scrutinize the fairness measures under consideration.

\begin{figure}[htb]
\centerline{
\includegraphics[width=0.88\textwidth]{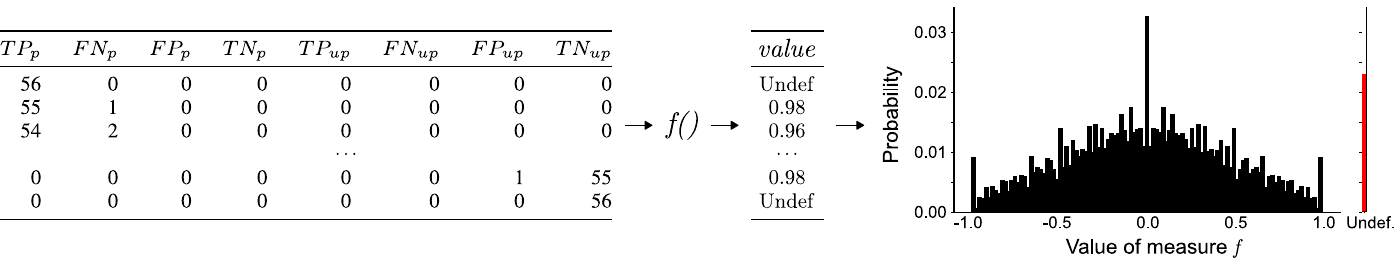}
}
\caption{Process of creating a histogram of a fairness measure probability mass function by generating all possible confusion matrices. Each table row represents an entry of a single confusion matrix (left). These confusion matrices map to fairness measure values (middle). By counting each measure value, we construct a histogram (right). The probability of an undefined measure value (Undef.) is represented as a separate red bar next to the histogram.} \label{fig:process}
\end{figure}

We will use probability mass functions depicted as histograms to analyze the effects different imbalance ratios and group ratios can have on the six fairness measures~(Eq.~\ref{eq:ae}--\ref{eq:npp}). In our analysis, we abstract from concrete classifiers or datasets and, therefore, assume that each possible confusion matrix is equally probable. We are aware that this is a strong assumption, however, in this way, we offer the most inclusive view of what may happen in any classification task. For example, although confusion matrices with mostly incorrect predictions seem less probable than those with mostly correct ones, classifiers tackling data streams will often temporarily make incorrect predictions due to concept drift~\cite{brzezinski2019dynamics}. Consequently, we will follow an approach similar to that presented in~\cite{brzezinski2019dynamics,DBLP:journals/isci/BrzezinskiSSS18} and generate all possible confusion matrices for a dataset size $n$, and calculate the measure's value for each matrix (Figure~\ref{fig:process}). Note that, for a dataset size $n$, the number of all possible confusion matrices defined by $k=8$ values ($\mathit{TP}_{p}, \mathit{FN}_{p}, \mathit{FP}_{p}, \mathit{TN}_{p}, \mathit{TP}_{up}, \mathit{FN}_{up}, \mathit{FP}_{up}, \mathit{TN}_{up}$) equals $c=\binom{n+k-1}{k-1}$. This formula is taken from the `stars and bars' theorem~\cite{feller}, which shows how to calculate the number of possible $k$-tuples of non-negative integers summing up to $n$.

Using the calculated measure values, we analyze the $\mathit{pmf}$-based histograms of each measure for varying $\mathit{IR}$ and $\mathit{GR}$. In our visualizations, we use $n=56$ with $\mathit{IR} \in \{ \frac{1}{28},\allowbreak \frac{1}{4},\allowbreak \frac{1}{2},\allowbreak \frac{3}{4},\allowbreak \frac{27}{28}\}$ and $\mathit{GR} \in \{ \frac{1}{28},\allowbreak \frac{1}{4},\allowbreak \frac{1}{2},\allowbreak \frac{3}{4},\allowbreak \frac{27}{28}\}$. We chose $n=56$ for two reasons: we wanted the value to be divisible by $k=8$ (the number of analyzed confusion matrix entries), and we needed to be able to compute all the possible $c=\binom{n+k-1}{k-1}$ confusion matrices using 32 GB of RAM. Although $n=56$ may seem like a small number, it corresponds to $c$=553,270,671 possible confusion matrices, which provides a respectable sample of classifier outputs.\footnote{It took 82 minutes to generate all the possible confusion matrices using an Apple M2 Pro with 32 GB of RAM.} The considered class and group proportions were selected to represent class balance ($\frac{1}{2}$), low imbalance ($\frac{1}{4}$ and $\frac{3}{4}$), and high imbalance ($\frac{1}{28}$ and $\frac{27}{28}$)~\cite{ratios}. Notice that the high imbalance ratios ($\frac{1}{28}$ and $\frac{27}{28}$) correspond to an extreme case where there are only two examples from the minority class, one for each protected/unprotected group.\footnote{This is the smallest number of examples needed to analyze binary class imbalance and fairness between two groups.} We also note that fairness measure values may be undefined for certain confusion matrices. This is the case when denominators in Eq.~\ref{eq:ae}--\ref{eq:npp} are equal to zero. The probability of obtaining such undefined values for a given $\mathit{IR}$/$\mathit{GR}$ will be visualized as a red bar on the right of the $\mathit{pmf}$ histogram (Figure~\ref{fig:process}).

\begin{figure}[h!]
\centerline{
    \includegraphics[width=0.96\textwidth]{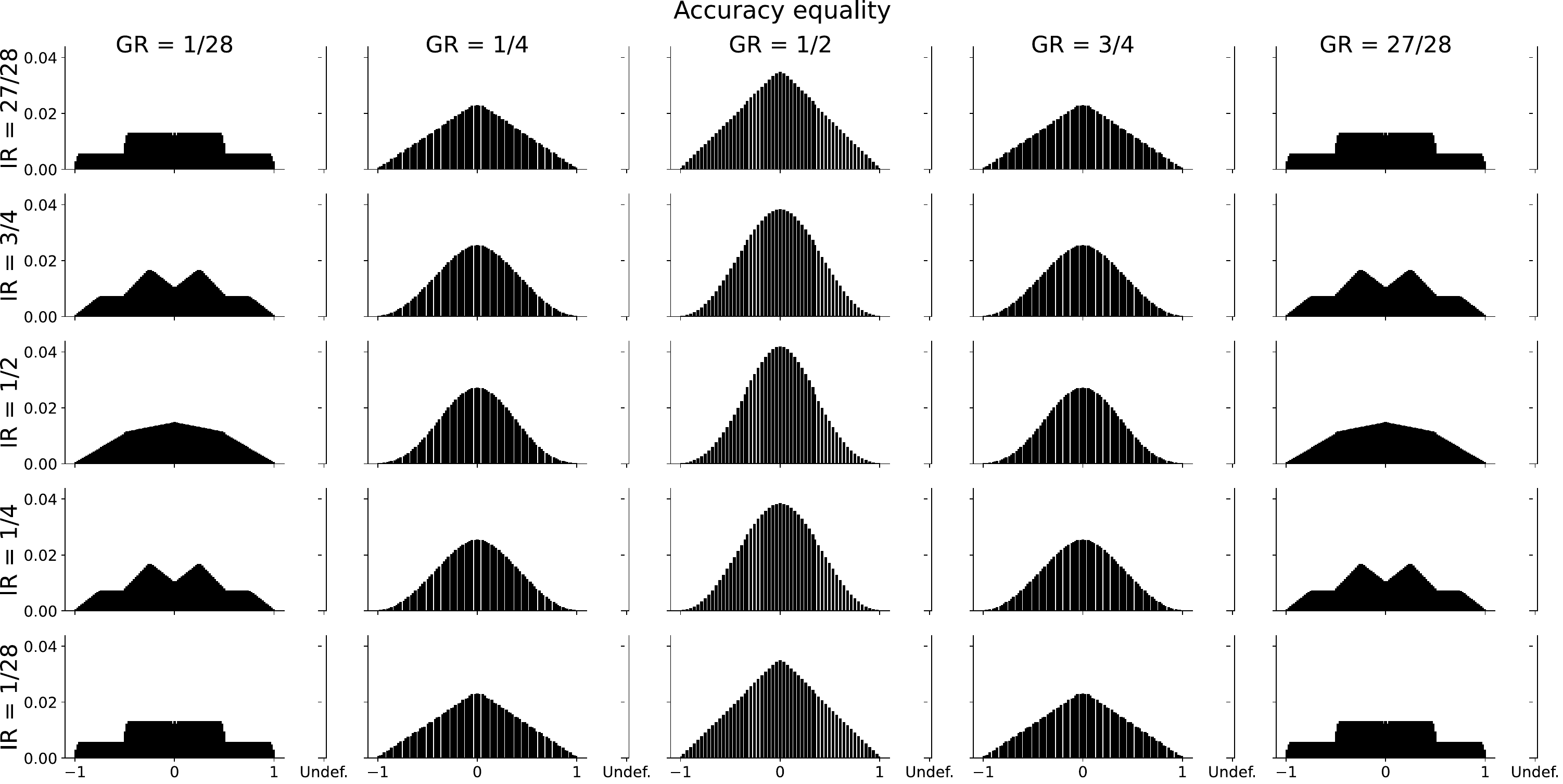}
}
\caption{Histograms of \textbf{accuracy equality} values for selected imbalance and group ratios. The x-axis shows possible measure values, whereas the y-axis shows the probability of obtaining a given value. Panels represent varying imbalance ratios $\mathit{IR}$ (top-bottom) and group ratios $\mathit{GR}$ (left-right). The probability of undefined values (Undef.) is represented as red bars next to each histogram.}\label{fig:hist_acc_eq}
\end{figure}

\begin{figure}
\centerline{
    \includegraphics[width=0.96\textwidth]{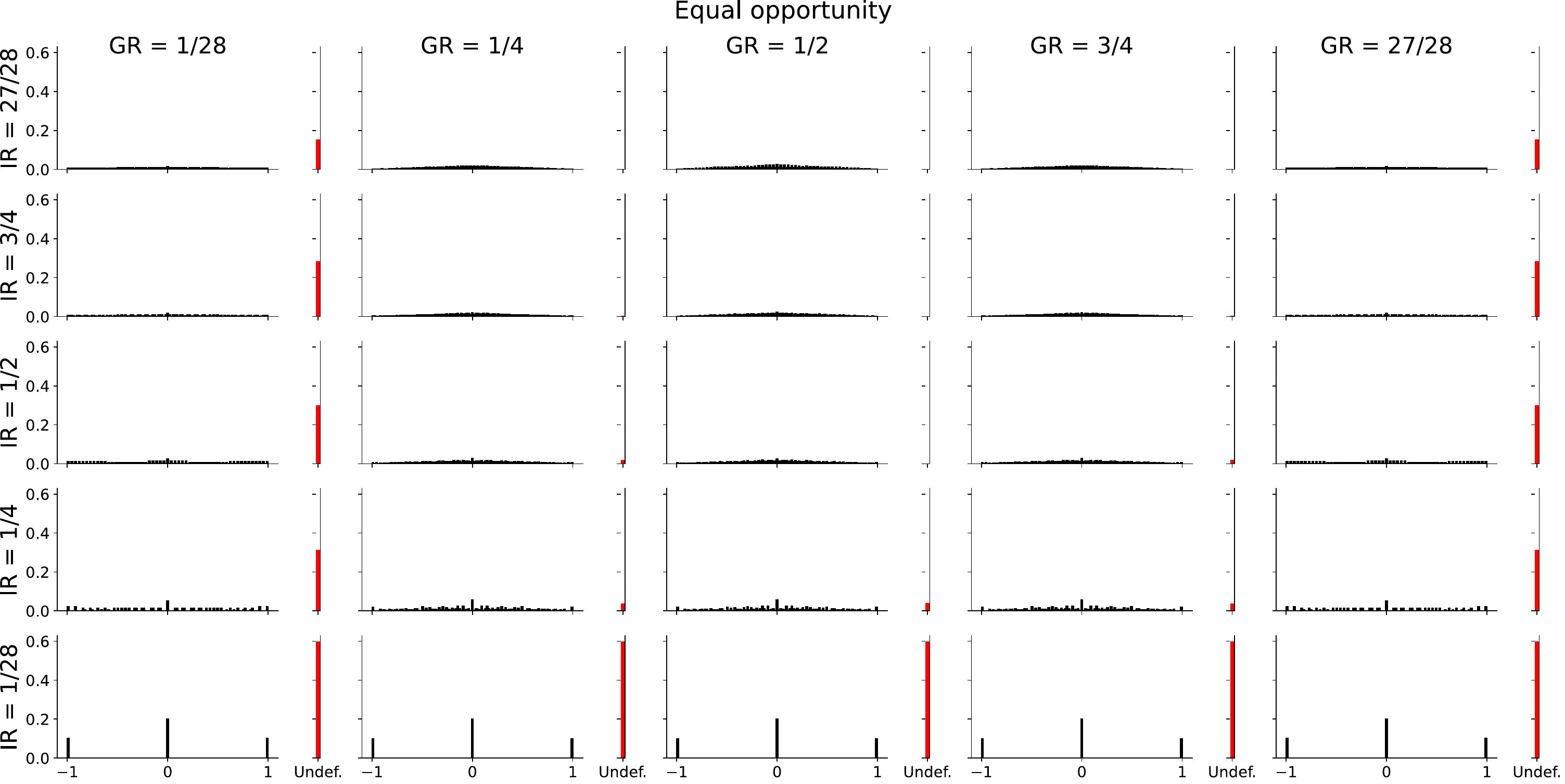}
}
\caption{Histograms of \textbf{equal opportunity} values for selected imbalance and group ratios. The x-axis shows possible measure values, whereas the y-axis shows the probability of obtaining a given value. Panels represent varying imbalance ratios $\mathit{IR}$ (top-bottom) and group ratios $\mathit{GR}$ (left-right). The probability of undefined values (Undef.) is represented as red bars next to each histogram.}\label{fig:hist_eq_op}
\end{figure}

\begin{figure}
\centerline{
    \includegraphics[width=0.96\textwidth]{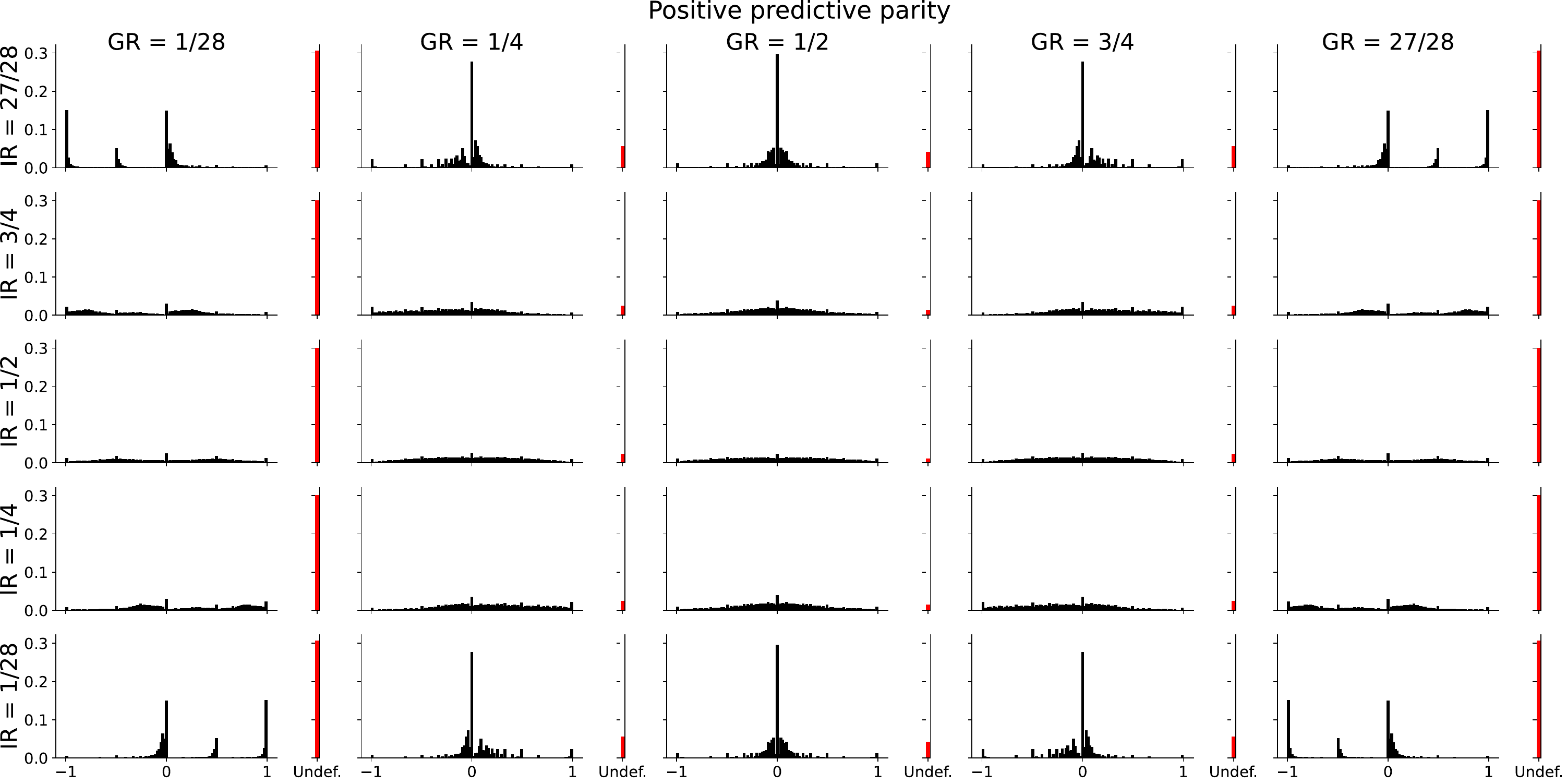}
}
\caption{Histograms of \textbf{positive predictive parity}  values for selected imbalance and group ratios. The x-axis shows possible measure values, whereas the y-axis shows the probability of obtaining a given value. Panels represent varying imbalance ratios $\mathit{IR}$ (top-bottom) and group ratios $\mathit{GR}$ (left-right). The probability of undefined values (Undef.) is represented as red bars next to each histogram.}\label{fig:hist_ppp}
\end{figure}

Figures~\ref{fig:hist_acc_eq}--\ref{fig:hist_ppp} present histograms for Accuracy Equality, Equal Opportunity, and Positive Predictive Parity. Statistical Parity (Supplementary Figure S2) has probability mass functions identical to Accuracy Equality, whereas Predictive Equality and Negative Predictive Parity are top-down mirror images of Equal Opportunity and Positive Predictive Parity, respectively (Supplementary Figures S4 and S6). The code for generating all the possible confusion matrices and their corresponding fairness measure values is available in the repository accompanying this paper.\footnote{Source codes
available at: \url{https://github.com/Rasalrai/analysis-of-fairness-measures/}.}

\clearpage

\begin{figure}[htb]
\centerline{
\includegraphics[width=0.44\textwidth]{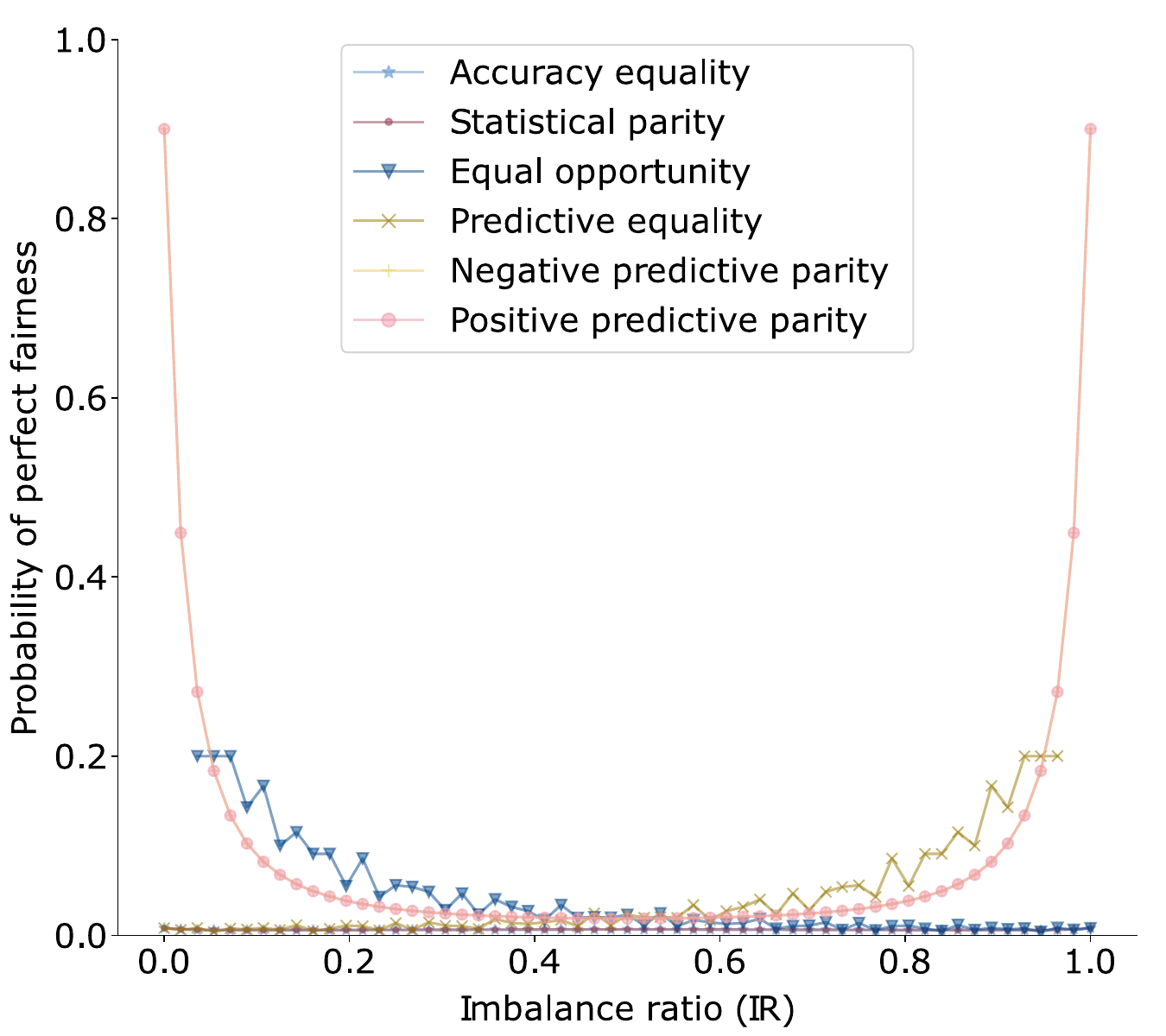}
\qquad
\includegraphics[width=0.44\textwidth]{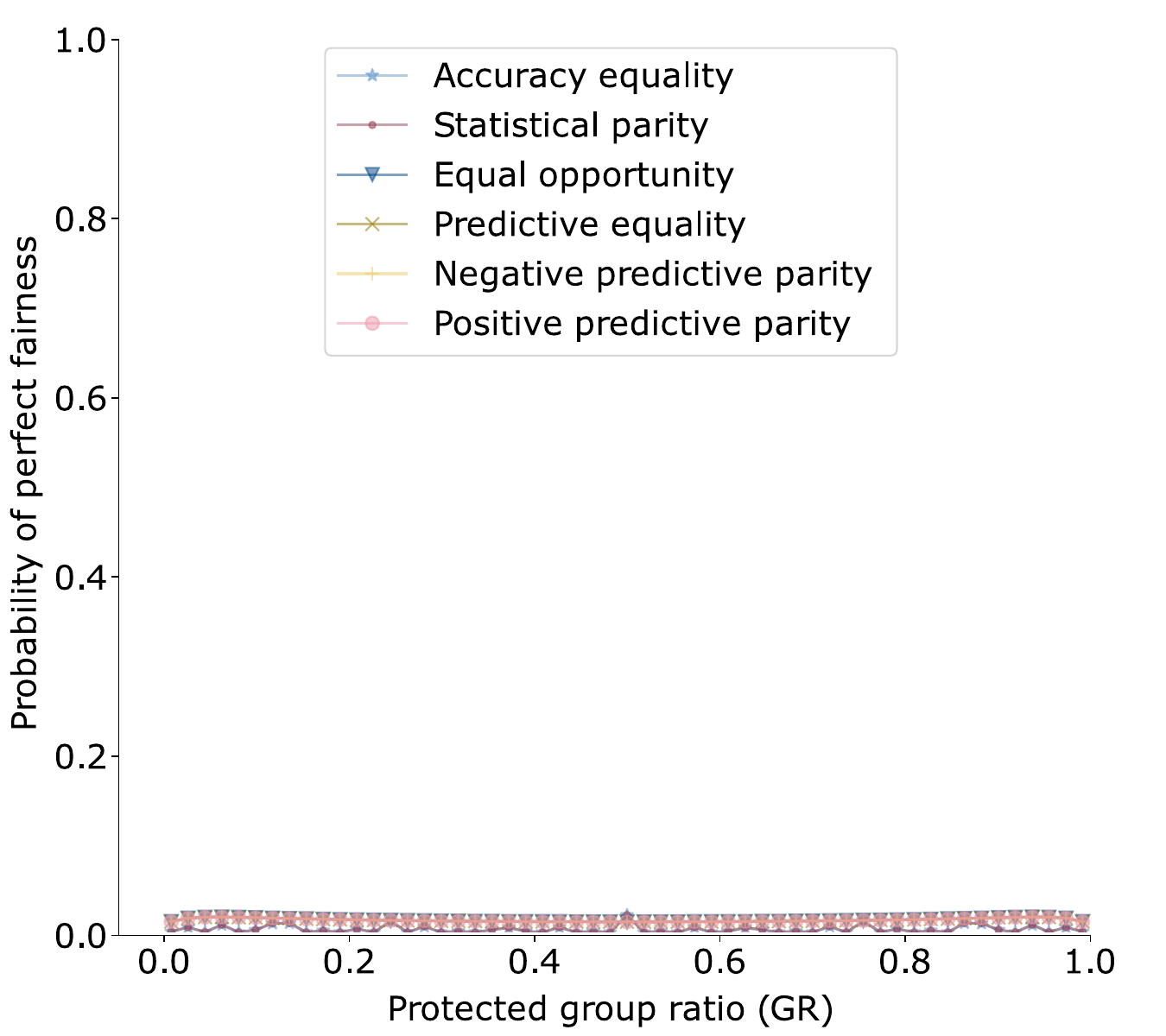}
}
\caption{Probability of achieving perfect fairness using different measures for varying class imbalance ratios (left) and varying group ratios (right).} \label{fig:perfect-fairness}%
\end{figure}

\begin{figure}[htb]
\centerline{
\includegraphics[width=0.44\textwidth]{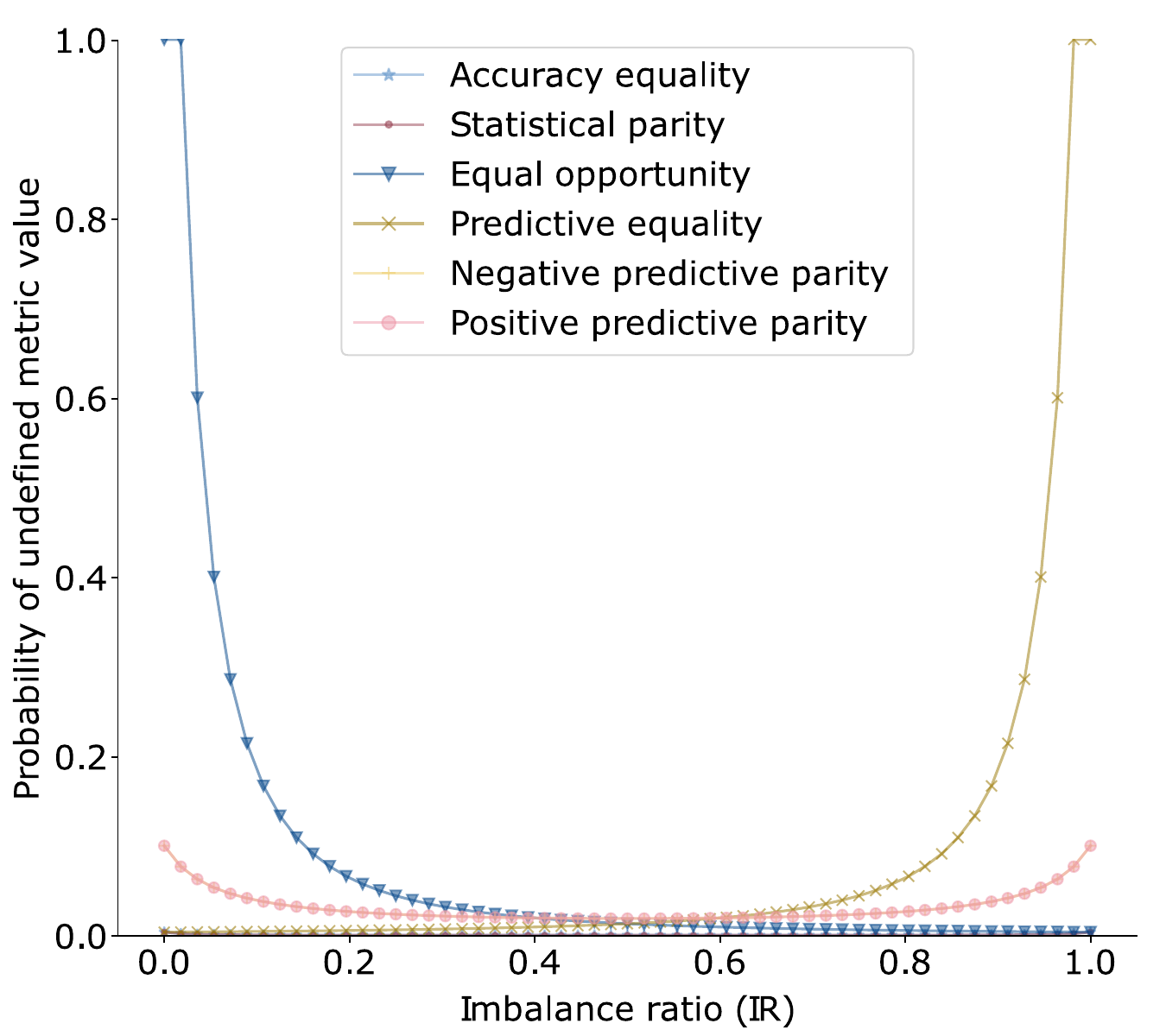}
\qquad
\includegraphics[width=0.44\textwidth]{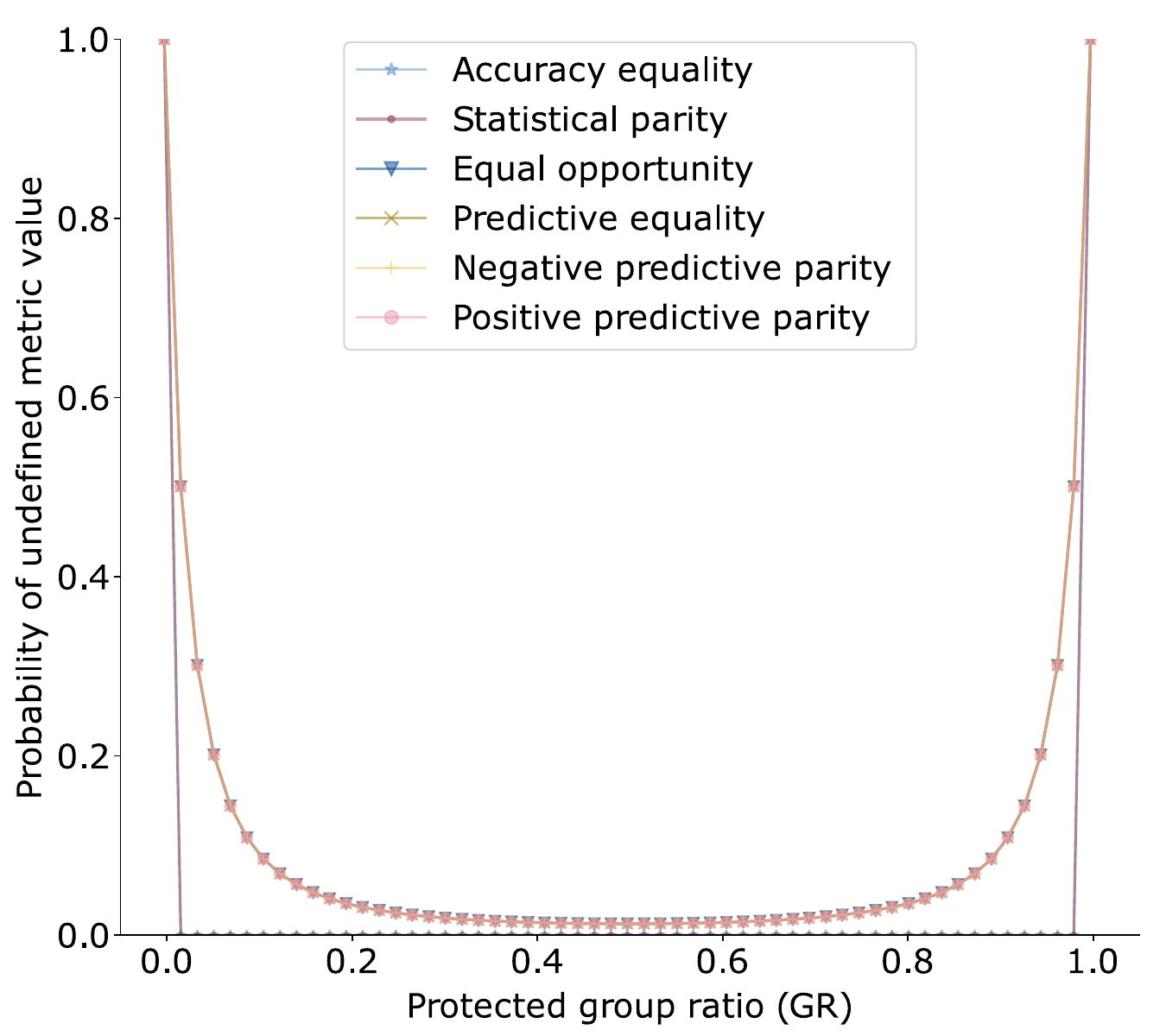}
}
\caption{Probability of achieving an undefined value for different measures under varying class imbalance ratios (left) and varying group ratios (right). Notice the abrupt growth in the number of undefined values for Accuracy Equality and Statistical Parity, and the gradual growth of the remaining four measures for varying group ratios (right).} \label{fig:nans}
\end{figure}

To complement the analysis, we also created line plots depicting the probability of achieving \textit{perfect fairness}, i.e., a fairness measure value equal to 0. Figure~\ref{fig:perfect-fairness} shows the fraction of confusion matrices corresponding to perfect fairness (y-axis) for varying imbalance and group ratios (x-axis). As can be noticed, extreme imbalance ratios can make it much more probable to achieve perfect fairness for some measures, whereas group ratio has a negligible effect (zoomed-in versions of the plots can be found in Supplementary Figures~S7 and S8). A similar plot was created to show the probabilities of obtaining undefined values of fairness measures (Figures~\ref{fig:nans}, S9, S10). Based on the presented histograms, perfect fairness plots, and undefined value plots, in the following section, we will analyze how the measure's probability mass functions change with varying $\mathit{IR}$ and $\mathit{GR}$ and propose a set of dataset-independent fairness measure properties.

\begin{figure}[htb]
\centerline{
\includegraphics[width=\textwidth]{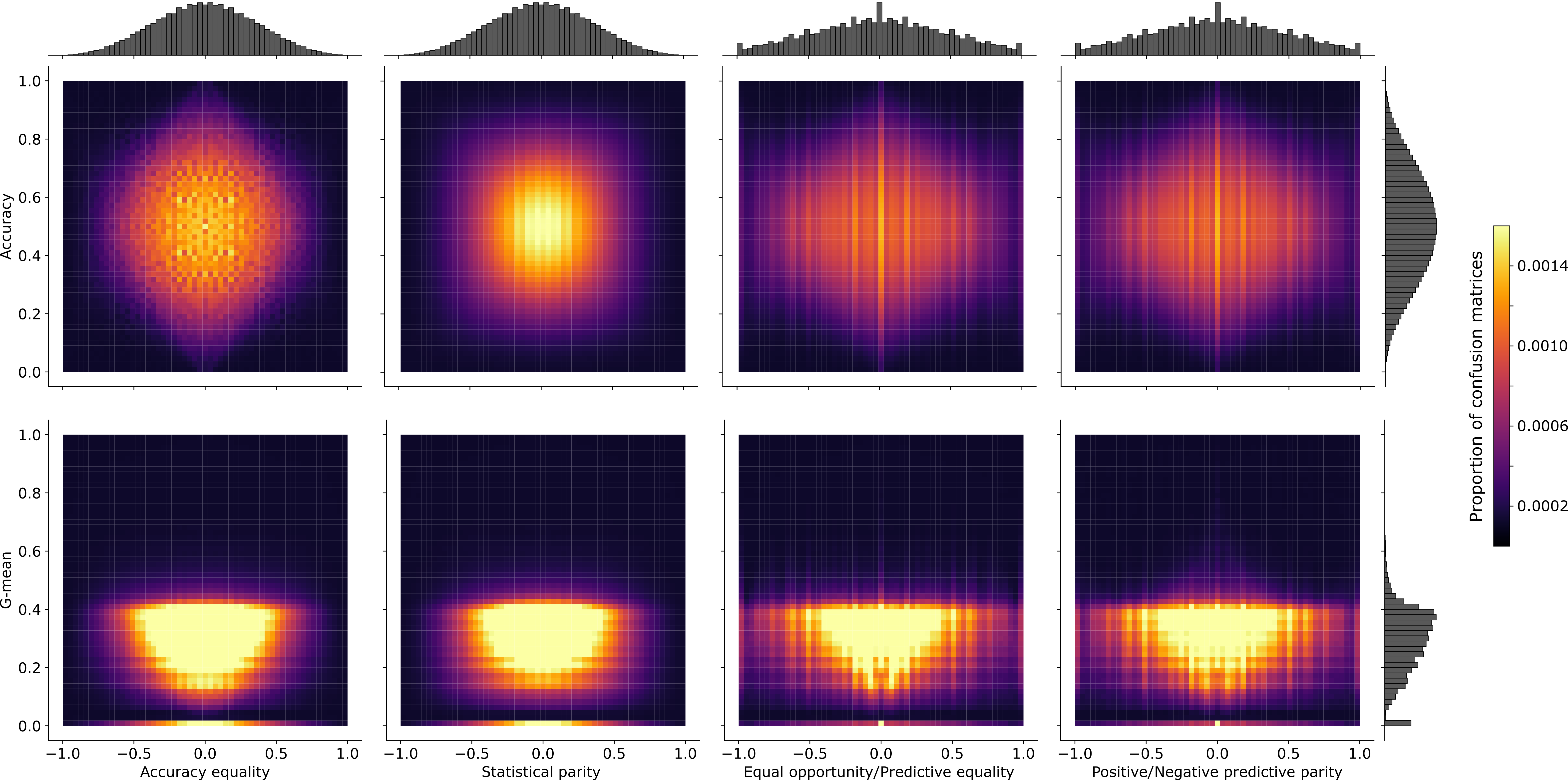}
}
\caption{Heatmaps showing the relation between different fairness measures (columns) and classification accuracy or G-mean (rows). Each heatmap shows the proportion of confusion matrices (color) that correspond to a given value of fairness (x-axis) and predictive performance (y-axis). Fairness measures with identical heatmaps have been grouped together.} \label{fig:heatmaps}
\end{figure}

Finally, the proposed method of analyzing fairness measures can also be applied to classification measures based on confusion matrices. In particular, one can analyze whether classifiers with high accuracy are less likely to be fair, simply due to the fact how fairness is measured. Figure~\ref{fig:heatmaps} tries to answer this question by showing the relation between different fairness measures (columns) and predictive performance measured by $\mathrm{Accuracy}=\frac{\mathit{TP}+\mathit{TN}}{n}$ and $\mathrm{G\mbox{-}mean}=\sqrt{\frac{\mathit{TP}}{\mathit{TP}+\mathit{FN}} \cdot \frac{\mathit{TN}}{\mathit{FP}+\mathit{TN}}}$ (rows). As can be noticed, perfect fairness can be achieved for any value of Accuracy or G-mean. Also, when values of Accuracy depart from 0.5, there is a larger proportion of confusion matrices corresponding to `close-to-ideal' than `unfair' confusion matrices, especially in the case of Equal Opportunity, Predictive Equality, Positive Predictive Parity, and Negative Predictive Parity  (the two rightmost panels in the first row).  Moreover,
the asymmetry in G-mean heatmaps (lower vs upper parts of all the panels in the second row) is due to the asymmetry of the G-mean distribution itself, which is much more likely to achieve lower values than higher ones. Additionally, as was seen in Figures~\ref{fig:hist_acc_eq}--\ref{fig:hist_ppp}, for particular high imbalance and group ratios the number of possible values of fairness can decrease rapidly. Therefore, although the analyzed definitions of fairness do not discriminate accurate classifiers in general, for datasets with particular example proportions oftentimes there is a fairness-accuracy tradeoff~\cite{Corbett-Davies_2017}.

\section{Properties of fairness measures}
\label{sec:properties}

With a method for analyzing fairness measures at hand, in this section, we will define and interpret the potentially desirable properties of these measures. The properties should aid researchers in comparing various measures, examining differences in their distributions, noticing unusual or unexpected values, and selecting measures suitable for a given classification task. Defining such a set of properties in the context of class imbalance and protected group bias is vital for several reasons. First, it helps understand the performance and limitations of these measures under varying dataset characteristics. As real-world datasets often have imbalanced classes and bias toward certain groups, understanding how these factors impact the fairness measures can guide their use. Second, defining these properties provides a theoretical framework that can help design novel fairness measures. Finally, it supports the development of more robust and fair machine learning models, as the proposed properties may help in the development of new fairness intervention methods for classifiers.

In this context, we postulate to analyze group fairness measures with respect to eight properties:
\begin{description}
    \item[Immunity to IR changes:] the distribution of the measure's values does not change for varying imbalance ratios. Measures with this property should perform the same way for datasets with different imbalance ratios.
    \item[Immunity to GR changes:] the distribution of the measure's values does not change for varying protected group ratios. Measures with this property should perform the same way for datasets with different group ratios.
    \item[Resolution Stability:] the number of unique measure values is large regardless of the imbalance and group ratios. Measures with this property will always be able to provide a wide range of different values, rather than degrading to signaling only a few values, e.g., perfect fairness (0) and perfect unfairness (1/-1).
    \item[Fairness Symmetry:] all the distributions of the measure values for varying $\mathit{IR}$ and $\mathit{GR}$ levels should be symmetrical around zero (i.e., around perfect fairness). Measures with this property will not promote one (protected/unprotected) group over the other for any class or group ratio.
    \item[IR Symmetry:] the measure's distribution is the same for counterpart ratios of positive and negative class examples (same distribution, e.g., for $\mathit{IR}=0.01$ and $\mathit{IR}=0.99$). Measures with this property focus on both classes and will behave the same way for a given proportion of classes, regardless of whether the positive or negative class is underrepresented.
    \item[GR Symmetry:] the measure's distribution is the same for counterpart ratios of protected and unprotected group examples (same distribution, e.g., for $\mathit{GR}=0.01$ and $\mathit{GR}=0.99$). Measures with this property focus on both protected and unprotected groups and will behave the same way for a given proportion of groups, regardless of whether the protected or unprotected group is underrepresented.
    \item[Perfect Fairness Stability:] the probability of achieving perfect fairness stays almost constant for different imbalance and group ratios. Measures with this property will make the task of achieving perfect fairness (e.g., by classifier interventions) comparable for different class and group proportions.
    \item[Undefined Values:] the existence of undefined values. Measures with fewer undefined values will make quantifying fairness less prone to numerical problems.
\end{description}

Having presented the visualization technique in Section~\ref{sec:distributions} and having defined the properties above, we will now use the proposed tools to analyze the fairness measures. Table~\ref{tab:properties} summarizes the results of the verification of the properties for each of the examined measures. Below we compare these outcomes, providing our observations.

\begin{table}
\begin{center}
{\caption{Properties of selected fairness measures; $^{\dagger}$: extreme group ratios introduce undefined values, but the shape of the distribution does not change.}\label{tab:properties}}
\begin{tabular}{lp{1.6cm}p{1.6cm}p{1.5cm}p{1.5cm}p{1.4cm}p{1.4cm}}
\toprule
 & {Accuracy Equality} & {Statistical Parity} & {Equal\qquad Opportunity} & {Predictive Equality} & {Positive Predictive Parity} & {Negative Predictive Parity} \\ \midrule
Immunity to IR changes & $\times$ & $\times$ & $\times$ & $\times$ & $\times$ & $\times$ \\
Immunity to GR changes & $\times$ & $\times$ & $\checkmark^\dagger$ & $\checkmark^\dagger$ & $\times$ & $\times$ \\
Resolution Stability & $\checkmark$ & $\checkmark$ & $\times$ & $\times$ & $\times$ & $\times$ \\
Fairness Symmetry & $\checkmark$ & $\checkmark$ & $\checkmark$ & $\checkmark$ & $\times$ & $\times$ \\
IR Symmetry & $\checkmark$ & $\checkmark$ & $\times$ & $\times$ & $\times$ & $\times$ \\
GR Symmetry & $\checkmark$ & $\checkmark$ & $\checkmark$ & $\checkmark$ & $\times$ & $\times$ \\
Perfect Fairness Stability & $\checkmark$ & $\checkmark$ & $\times$ & $\times$ & $\times$ & $\times$ \\
Undefined Values & when $n_p=0$ or $n_{\mathit{up}}=0$ & when $n_p=0$ or $n_{\mathit{up}}=0$ & low/high GR, low IR & low/high GR, high IR & low/high GR & low/high GR \\ \bottomrule
\end{tabular}
\end{center}
\end{table}

Let us start by noticing that the six analyzed measures form three groups: i) Accuracy Equality and Statistical Parity, ii) Equal Opportunity and Predictive Equality, and iii) Positive Predictive Parity and Negative Predictive Parity. Indeed, the measures within the pairs have similar definitions (see Eq. ~\ref{eq:ae}--\ref{eq:npp}), e.g., the $\mathit{TN}_{p}$ and $\mathit{TN}_{up}$ in Accuracy Equality are simply substituted by $\mathit{FP}_{p}$ and $\mathit{FP}_{up}$ in Statistical Parity. Additionally, these measure pairs either have the same value distributions (see, e.g.,  Figure~\ref{fig:hist_acc_eq} and Figure~S2 in the supplementary materials) or their distributions are symmetrical, e.g., Equal Opportunity focuses on the positive class, whereas Predictive Equality focuses on the negative class (see Figure~\ref{fig:hist_eq_op} and Figure~S4 in the supplementary materials). The same can be noticed on Figures~\ref{fig:perfect-fairness} and~\ref{fig:nans}. As a result, the columns in Table~\ref{tab:properties} representing the measures from each pair have identical entries.

Now, let us consider the first three properties. It can be noticed that each measure is susceptible to changes in class and group ratios. For varying imbalance ratios (examine the panels in Figures~\ref{fig:hist_acc_eq}--\ref{fig:hist_ppp} from top to bottom), each measure changes its distribution, with Equal Opportunity/Predictive Equality having very few unique values (low resolution) for low/high $\mathit{IR}$, and Positive and Negative Predictive Parity also having fewer unique values and much bigger chances of perfect fairness for extreme imbalance ratios (see also Figure~\ref{fig:perfect-fairness}). Although Accuracy Equality and Statistical Parity also change their probability mass functions with class imbalance, the differences are much less drastic and do not affect the number of unique measure values. Similarly, for varying group ratios (examine the panels in Figures~\ref{fig:hist_acc_eq}--\ref{fig:hist_ppp} from left to right), the measures change their distributions, but these changes do not influence the chances of achieving perfect fairness that much, with Equal Opportunity and Predictive Equality practically having the same distributions, only with a higher chance of obtaining an undefined value (Figures~\ref{fig:perfect-fairness} and~\ref{fig:nans}).

The next three properties ascertain the symmetry of measure behavior. Practically all measure distributions (Figures~\ref{fig:hist_acc_eq}-\ref{fig:hist_ppp}) are always centered around zero, i.e. perfect fairness. Only Positive and Negative Predictive Parity become asymmetrical for imbalanced group proportions. Moreover, Accuracy Equality and Statistical Parity additionally have the same distributions for counterpart group and class proportions, making these two measures the most symmetrical in terms of treating positive/negative and protected/unprotected examples. On the other hand, Equal Opportunity and Predictive Equality are only symmetrical in treating protected/unprotected groups, whereas Positive and Negative Predictive Parity favor protected or unprotected groups depending on $\mathit{IR}$ and $\mathit{GR}$.

Taking a closer look at the chances of achieving perfect fairness (Figure~\ref{fig:perfect-fairness}), we see that only Accuracy Equality and Statistical Parity offer consistency.  The remaining measures, on the other hand, have substantially higher theoretical chances of signaling perfect fairness depending on the imbalance ratio and group ratio. In particular, it is interesting to note that for Equal Opportunity, there is a bigger chance of achieving perfect fairness when there are few positive examples in the dataset. In contrast, Predictive Equality makes it easier to achieve perfect fairness when there are few negative examples.

Finally, the chances of obtaining an undefined value differ between measures (Figure~\ref{fig:nans}). We see that only Accuracy Equality and Statistical Parity are almost free of undefined values---they only occur when one of the groups is missing ($n_p=0$ or $n_{\mathit{up}}=0$). The remaining measures, on the other hand, can have high probabilities of undefined values depending on the imbalance ratio and group ratio. In particular, for very high and very low values of $\mathit{GR}$, undefined values are the dominant values for Equal Opportunity, Predictive Equality, Positive Predictive Parity, and Negative Predictive Parity (Figures~\ref{fig:hist_eq_op} and \ref{fig:hist_ppp}). Additionally, $\mathit{IR}$ has a strong effect on the number of undefined values for Equal Opportunity and Predictive Equality. Although undefined values can be considered merely a numerical problem, not being able to measure fairness can impact the ability to act upon unfair predictions.

In the following section, we will verify whether these dataset-independent properties can affect the measured fairness of machine learning models in a practical classification scenario.

\section{Case study on the effect of measure properties on classifier fairness}
\label{sec:case-study}

\subsection{Experimental setup}
To verify how the proposed properties apply to practical problems involving real classifiers, we performed an experiment using the UCI Adult dataset~\cite{Dua2019dataset}. We chose UCI Adult as it is one of the most popular binary datasets in the context of both imbalanced learning and model fairness~\cite{fabris2022fairness}. The prediction task is binary classification: the positive class `$>50K$' (also referred to as the \textit{rich}) indicates people with over 50,000 USD of yearly income, and the negative class `$<=50K$' (\textit{poor}) denotes a yearly income not higher than 50,000 USD. We selected \textit{sex} as the protected attribute, with \textit{Female} considered as the protected group and \textit{Male} as the unprotected group.

Using the UCI Adult dataset, our goal was to see whether the fairness of classifiers (as quantified by different measures) will be affected by different levels of protected group ratios ($\mathit{GR}$) and imbalance ratios ($\mathit{IR}$). Therefore, we decided to sample subsets from this dataset with controlled values of $\mathit{IR}$ and $\mathit{GR}$. Each subset had a specified value for one of the ratios (the selected ratios $\{0.01,\allowbreak0.02,\allowbreak0.05,\allowbreak0.1,\allowbreak0.2,\allowbreak0.3,\allowbreak0.4,\allowbreak0.5,\allowbreak0.6,\allowbreak0.7,\allowbreak0.8,\allowbreak0.9,\allowbreak0.95,\allowbreak0.98,\allowbreak0.99\}$), while the other ratio was set to 0.5. All the sampled subsets of the dataset were equally sized to $n=1100$ in order to avoid the impact of differently-sized data on the results. Special care was taken to ensure that the proportions of groups were in accordance with the given $\mathit{GR}$ within the entire dataset as well as within each class. For instance, for a data subset with $\mathit{IR} = 0.1$ and $\mathit{GR} = 0.5$, there were 55 poor women, 55 rich women, 495 poor men, and 495 rich men in this subset. All these conditions were set to make the classification problems for each data subset as similar as possible, with only $\mathit{IR}$ or only $\mathit{GR}$ changing.

Using the prepared data subsets with given $\mathit{IR}$ and $\mathit{GR}$, for each subset, we performed 50 repetitions of randomly stratified holdout evaluations (67\% train, 33\% test). In each evaluation, we assessed the fairness of six types of popular learning algorithms chosen for their diversity: k-Nearest Neighbors (k-NN), Naive Bayes, Decision Tree, Logistic Regression, Random Forest, a Multilayer Perceptron with a hidden layer of 100 neurons (MLP). Since our goal is only to illustrate the effect of varying imbalance and group ratios, we left the classifiers with default parameters in their Python implementation in the scikit-learn library~\cite{scikit-learn}. The experiments were conducted on a machine equipped with Intel\copyright \ Core\texttrademark \ i7-1260P 4.7GHz processor and 48 GB of RAM, using Python 3.10.10 and scikit-learn version 1.2.2. Reproducible scripts for data preparation and all experiments are available at: \url{https://github.com/Rasalrai/analysis-of-fairness-measures/}.

\subsection{Results}
\begin{figure}[htb]
\centerline{
    \includegraphics[width=0.9\textwidth]{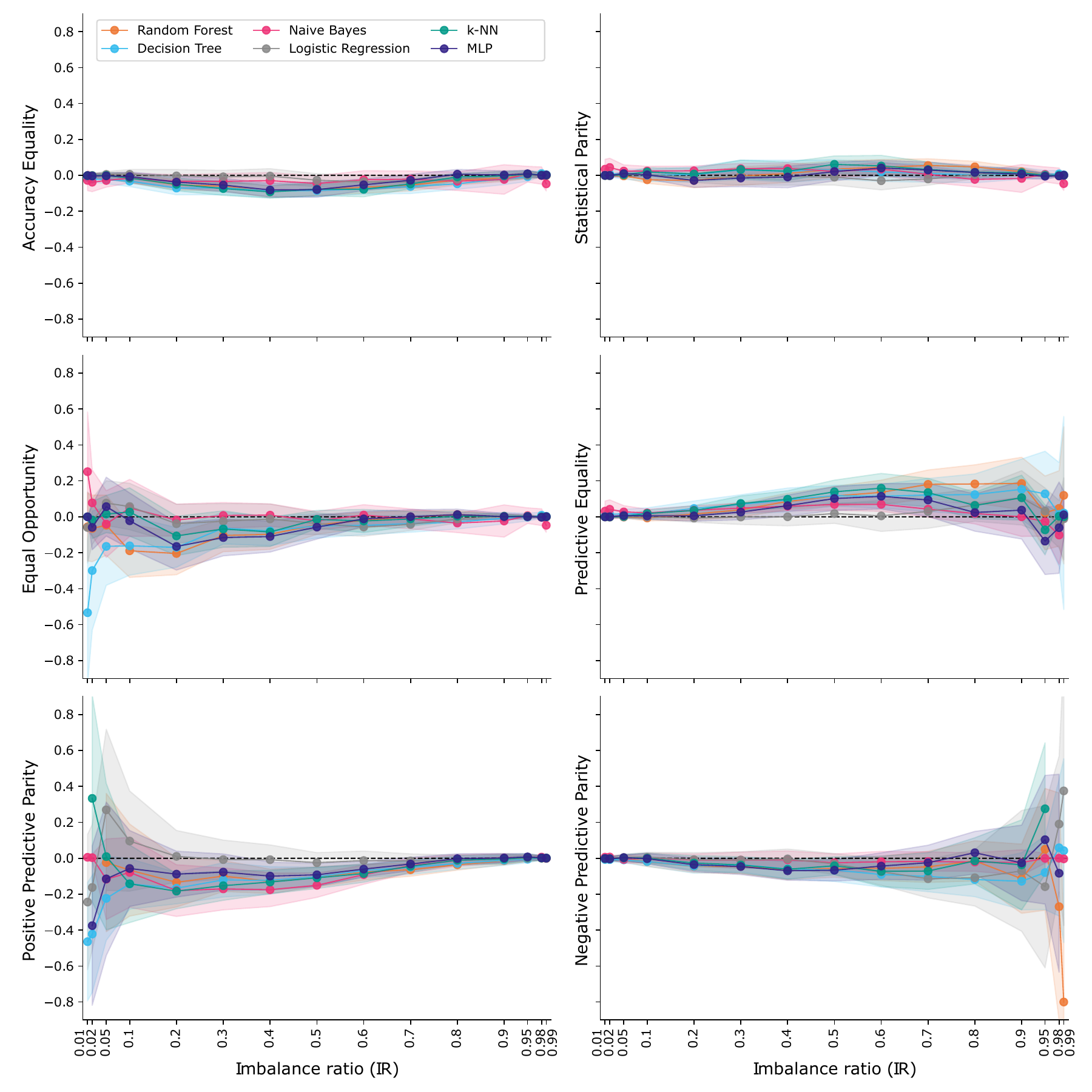}
}
\caption{Fairness values achieved by the analyzed classifiers for different fairness measures, as a function of imbalance ratio $\mathit{IR}$ with $\mathit{GR}$ set to 0.5. Points represent mean classifier fairness for the 50 repetitions for the given $\mathit{IR}$ and the filled area shows the standard deviation.} \label{fig:case_study_graph}
\end{figure}

Figure~\ref{fig:case_study_graph} presents the means and standard deviations (y-axis) of the analyzed fairness measures for varying imbalance ratios $\mathit{IR}$ (x-axis). An analogous plot for varying group ratios $\mathit{GR}$ can be found in Supplementary Figure S8.

As the results show, the fairness of the classifiers depends on the measure used and the imbalance ratio in the dataset. The most stable fairness is achieved according to Accuracy Equality and Statistical Parity. These results are in line with our histogram-based analysis and the properties assigned to these two measures (Table~\ref{tab:properties}). Moreover, one can notice that fairness assessments according to Equal Opportunity and Predictive Equality become unstable for imbalance ratios corresponding to very few unique values (Resolution Stability property). Therefore, Equal Opportunity may be less reliable for low values of $\mathit{IR}$ (few positives), whereas Predictive Equality might not be the best choice for datasets with high values of $\mathit{IR}$ (few negatives). Finally, Positive and Negative Predictive Parity are even more unstable than the previous two measures. This is the combined effect of very few unique values (Resolution Stability) and the measures relying on predicted positives ($\mathit{FP}$ and $\mathit{TP}$) or predicted negatives ($\mathit{FN}$ and $\mathit{TN}$) rather than actual positives and negatives as was the case for Equal Opportunity and Predictive Equality. Therefore, the measures are more prone to biased classifier predictions. For varying $\mathit{GR}$, all the measures performed very similarly (Supplementary Figure S12). Indeed, as it was mentioned in Section~\ref{sec:properties}, the analyzed measures are much less prone to changes in protected group ratios.

\section{Discussion}
\label{sec:discussion}
The presented study shows that data imbalance can have a non-negligible effect on the behavior of group fairness measures. In particular, positive predictive parity and negative predictive parity behave asymmetrically depending on the imbalance ratio, which makes it easier to achieve high fairness using these measures simply due to class imbalance. Predictive equality and equal opportunity showcase similar problems, albeit to a lesser extent. This highlights the need for careful selection of fairness measures depending on the analyzed data and the potential need for new measures that are more immune to class imbalance.

Apart from restating the main findings of this study, it is worth listing its limitations. First, we note that the presented analysis focused on group fairness measures defined on the basis of entries of confusion matrices. Several other measures quantify algorithmic fairness through a different perspective, e.g., individual fairness~\cite{caton2020fairness}. To study the effect of class imbalance on those measures, methods different than the one proposed in this paper are required. Similarly, we focused on problems with two classes and two groups. The analysis of multiple classes and multiple groups using our approach would be possible but would require analyzing the data using a one-vs-all strategy. Also, even though the number of simulated confusion matrices is very large ($c$=553,270,671), the fact that the sum of the entries of these confusion matrices is $n=56$ may lead to an exaggeration of undefined and perfect fairness situations compared to datasets with much larger $n$. That being said, all the properties in Table~\ref{tab:properties} hold true regardless of the dataset size. Finally, the experimental case study uses only one popular fairness benchmark, which has its limitations~\cite{NEURIPS2021_32e54441}. Therefore, more real-world experiments, in particular in the domains of image and text analysis, are still required to assess the practical impact of the characteristics of fairness measures discussed in this paper. Nevertheless, the properties defined in Section~\ref{sec:properties} are independent of the dataset and classifier type and can be supportive for practitioners regardless of their problem setting.

Our claim that the discussed properties of fairness measures are independent of any dataset and classifier relies on the fact that we exhaustively analyze all possible confusion matrices. However, one may argue that not all confusion matrices are equally probable. Indeed, classifiers that perform worse than a majority stub are not likely to be used in practice~\cite{Fairea_2021}. Nevertheless, we decided to analyze all possible confusion matrices because this approach relies on fewer assumptions. What constitutes a `good enough' classifier differs from application to application. Especially in the domain of class-imbalanced learning, practitioners are willing to trade overall accuracy in favor of correct detection of the minority class, often expressed by means of specialized measures of predictive performance~\cite{he2009learning,gu2009evaluation}. Moreover, as mentioned earlier, classifiers tackling concept-drifting data streams will have periods of incorrect predictions after sudden changes in the data generating distribution~\cite{brzezinski2019dynamics}. For these reasons, we decided to treat each confusion matrix as equally probable; nevertheless, future studies may focus on more specific cases with more assumptions about the data distribution and classifier performance.

\section{Conclusions}
\label{sec:conclusions}
In this paper, we have analyzed the behavior of six popular group fairness measures in the context of varying class imbalance and protected group bias. For this purpose, we have defined eight dataset-independent properties that helped us characterize the studied measures using their probability mass functions. We further verified the proposed general properties through a controlled experiment using real-world data and six different classifiers.

Our results show that all the analyzed measures change their behavior in the presence of class imbalance and, to a lesser extent, in the presence of protected group bias. In particular, we have shown that measures that take into account the entire confusion matrix, such as Accuracy Equality and Statistical Parity, have the most stable value distributions under varying class and group proportions, treat both classes and groups symmetrically, have hardly any undefined values, and the chance of achieving perfect fairness stays close to constant for all imbalance and group ratios. Therefore, these measures can be considered the most reliable for imbalanced datasets. We have also highlighted that Equal Opportunity and Predictive Equality are complementary, as the first one becomes less stable with few positive examples in a dataset, and the latter performs worse with few negative examples in a dataset. Therefore, depending on the type of class imbalance, one will work better than the other. Finally, Positive and Negative Predictive Parity were found to be the least stable and most asymmetric in their distributions. That is why these two measures should be used mainly for datasets with relatively balanced classes and protected groups. 

The findings of this study can be directly used to improve the fairness of machine learning models in two ways. First, our study highlights the need to select fairness measures while taking into account data characteristics. Secondly, we show which measures are more suitable for which types of datasets, guiding fairness measure selection in practical settings. Moreover, our study opens several avenues for further research. Future work could extend the current analysis to properties of fairness measures under more complex data scenarios. In particular, we would be interested in investigating other types of uneven data distributions, such as stereotypical bias. The term stereotypical bias refers to the situation when protected groups are underrepresented within certain classes, even though they are evenly distributed in the entire dataset. Recent works have shown that, indeed, stereotypical bias occurs in popular computer vision datasets~\cite{dominguez2023gender}. Moreover, one potential limitation of this study is the assumption that all possible confusion matrices are equally probable. Further studies could implement the probability of achieving particular confusion matrices as a parameter that could be used to analyze the properties of measures under different priors. Additionally, the properties put forward in this paper could be used to design novel fairness measures that are more robust and better suited to handle varying class imbalance and protected group bias. Finally, these properties may aid in the development of new fairness intervention methods for classifiers, ultimately contributing to the creation of more responsible AI systems.

\begin{acks}
This research was partly funded by the National Science Centre, Poland, grant number 2022/47/D/ST6/01770. For the purpose of Open Access, the author has applied a CC-BY public copyright license to any Author Accepted Manuscript (AAM) version arising from this submission.
\end{acks}

\bibliographystyle{ACM-Reference-Format}
\bibliography{fairness}

\clearpage

\appendix
\renewcommand{\theequation}{S\arabic{equation}}
\renewcommand{\thefigure}{S\arabic{figure}}
\renewcommand{\thetable}{S\arabic{table}}
\renewcommand{\thesection}{S\arabic{section}}
\setcounter{figure}{0}    
\setcounter{table}{0}  

\section*{Supplement}

\begin{figure}[h!]
\centerline{
    \includegraphics[width=0.95\textwidth]{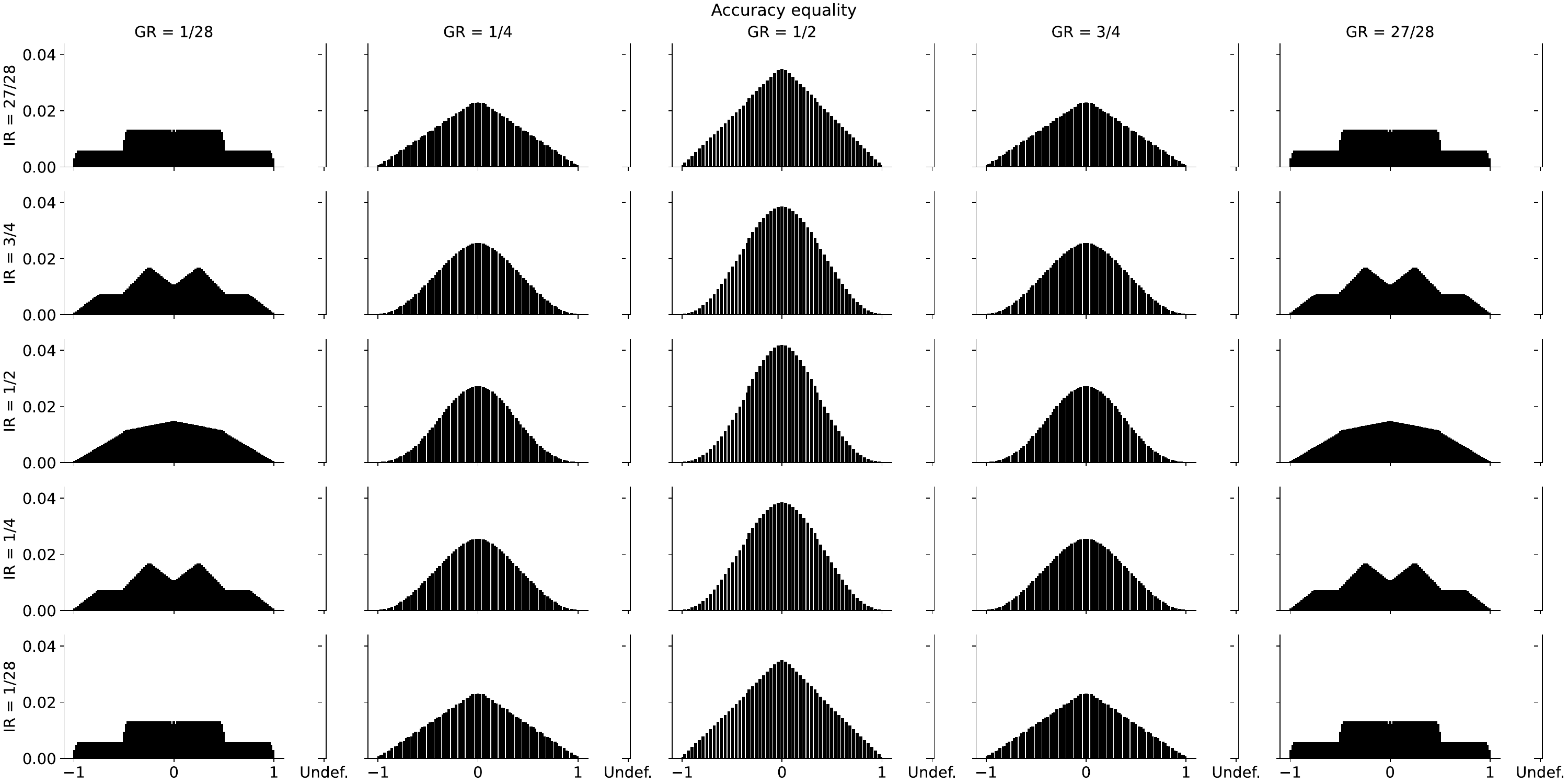}
}
\caption{Histograms of \textbf{accuracy equality} values for selected  imbalance and group ratios. The x-axis spans between the minimum and maximum of each measure ($[-1, 1]$) and is divided into bins of equal width. The y-axis shows the probability of obtaining a given measure value. For a given measure, different panels represent varying imbalance ratios $\mathit{IR}$ (top-bottom) and group ratios $\mathit{GR}$ (left-right). The probability of undefined values (Undef.) is represented as red bars next to each histogram.
}\label{fig:hist_stat_parity}
\end{figure}

\begin{figure}
\centerline{
    \includegraphics[width=0.95\textwidth]{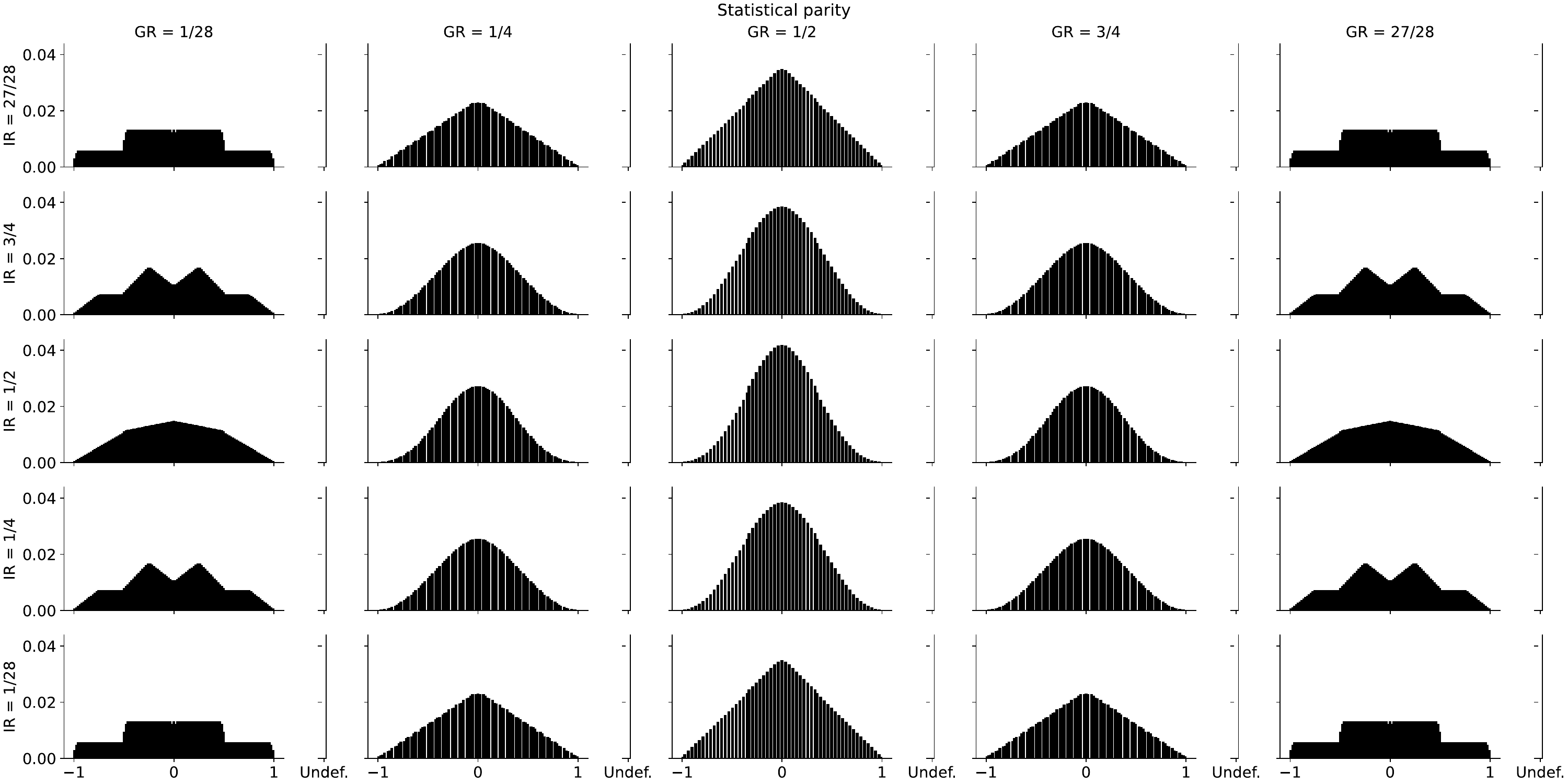}
}
\caption{Histograms of \textbf{statistical parity} values for selected  imbalance and group ratios. The x-axis spans between the minimum and maximum of each measure ($[-1, 1]$) and is divided into bins of equal width. The y-axis shows the probability of obtaining a given measure value. For a given measure, different panels represent varying imbalance ratios $\mathit{IR}$ (top-bottom) and group ratios $\mathit{GR}$ (left-right). The probability of undefined values (Undef.) is represented as red bars next to each histogram.}\label{fig:hist_acc_equality}
\end{figure}

\begin{figure*}
\centerline{
    \includegraphics[width=0.95\textwidth]{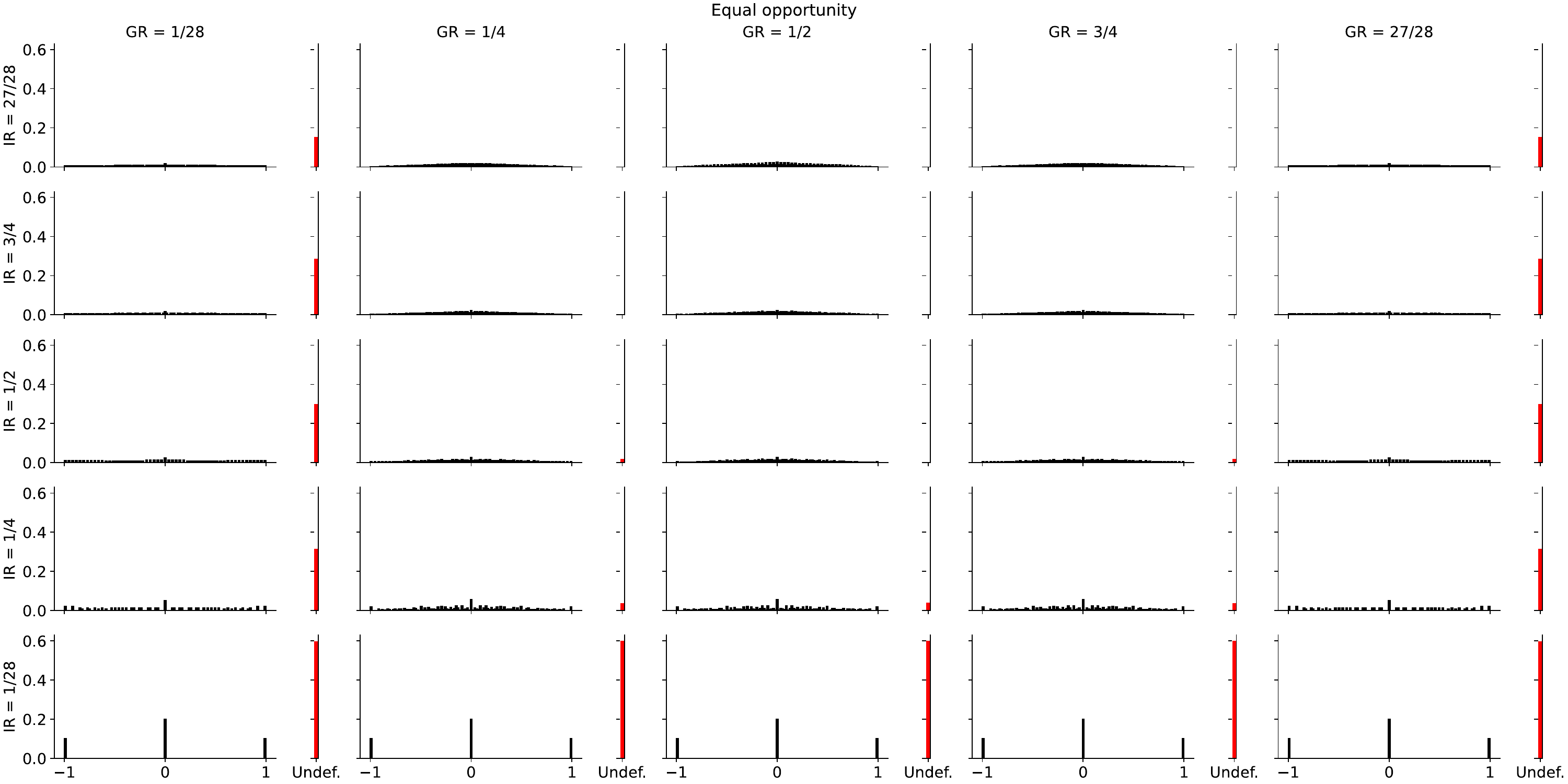}
}
\caption{Histograms of \textbf{equal opportunity} values for selected  imbalance and group ratios. The x-axis spans between the minimum and maximum of each measure ($[-1, 1]$) and is divided into bins of equal width. The y-axis shows the probability of obtaining a given measure value. For a given measure, different panels represent varying imbalance ratios $\mathit{IR}$ (top-bottom) and group ratios $\mathit{GR}$ (left-right). The probability of undefined values (Undef.) is represented as red bars next to each histogram.}\label{fig:hist_eq_opportunity}
\end{figure*}

\begin{figure*}[h!]
\centerline{
    \includegraphics[width=0.95\textwidth]{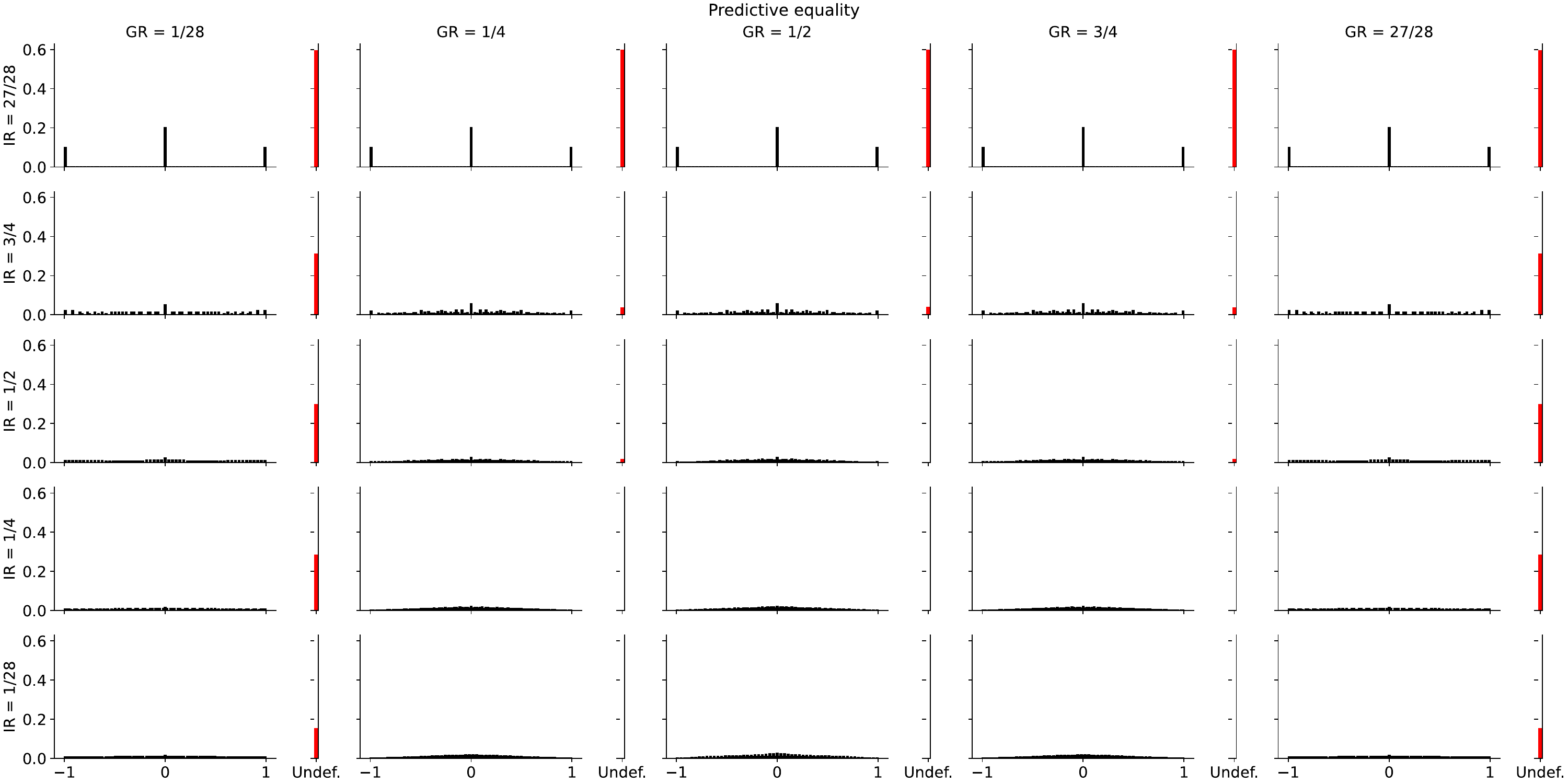}
}
\caption{Histograms of \textbf{predictive equality} values for selected  imbalance and group ratios. The x-axis spans between the minimum and maximum of each measure ($[-1, 1]$) and is divided into bins of equal width. The y-axis shows the probability of obtaining a given measure value. For a given measure, different panels represent varying imbalance ratios $\mathit{IR}$ (top-bottom) and group ratios $\mathit{GR}$ (left-right). The probability of undefined values (Undef.) is represented as red bars next to each histogram.}\label{fig:hist_pred_equality}
\end{figure*}

\begin{figure*}[h!]
\centerline{
    \includegraphics[width=0.95\textwidth]{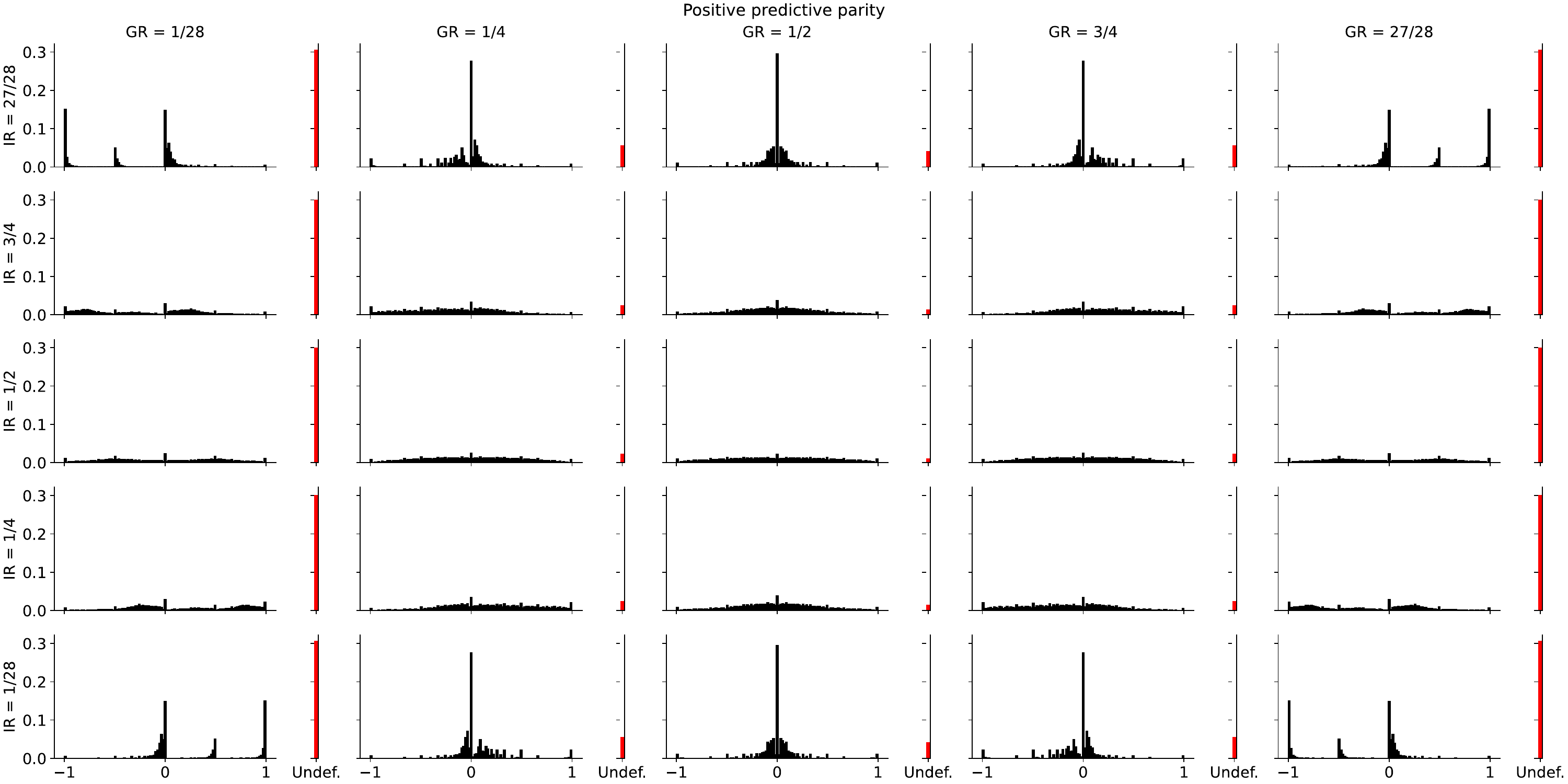}
}
\caption{Histograms of \textbf{positive predictive parity} values for selected  imbalance and group ratios. The x-axis spans between the minimum and maximum of each measure ($[-1, 1]$) and is divided into bins of equal width. The y-axis shows the probability of obtaining a given measure value. For a given measure, different panels represent varying imbalance ratios $\mathit{IR}$ (top-bottom) and group ratios $\mathit{GR}$ (left-right). The probability of undefined values (Undef.) is represented as red bars next to each histogram.}\label{fig:hist_pos_pred_parity}
\end{figure*}

\begin{figure*}[h!]
\centerline{
    \includegraphics[width=0.95\textwidth]{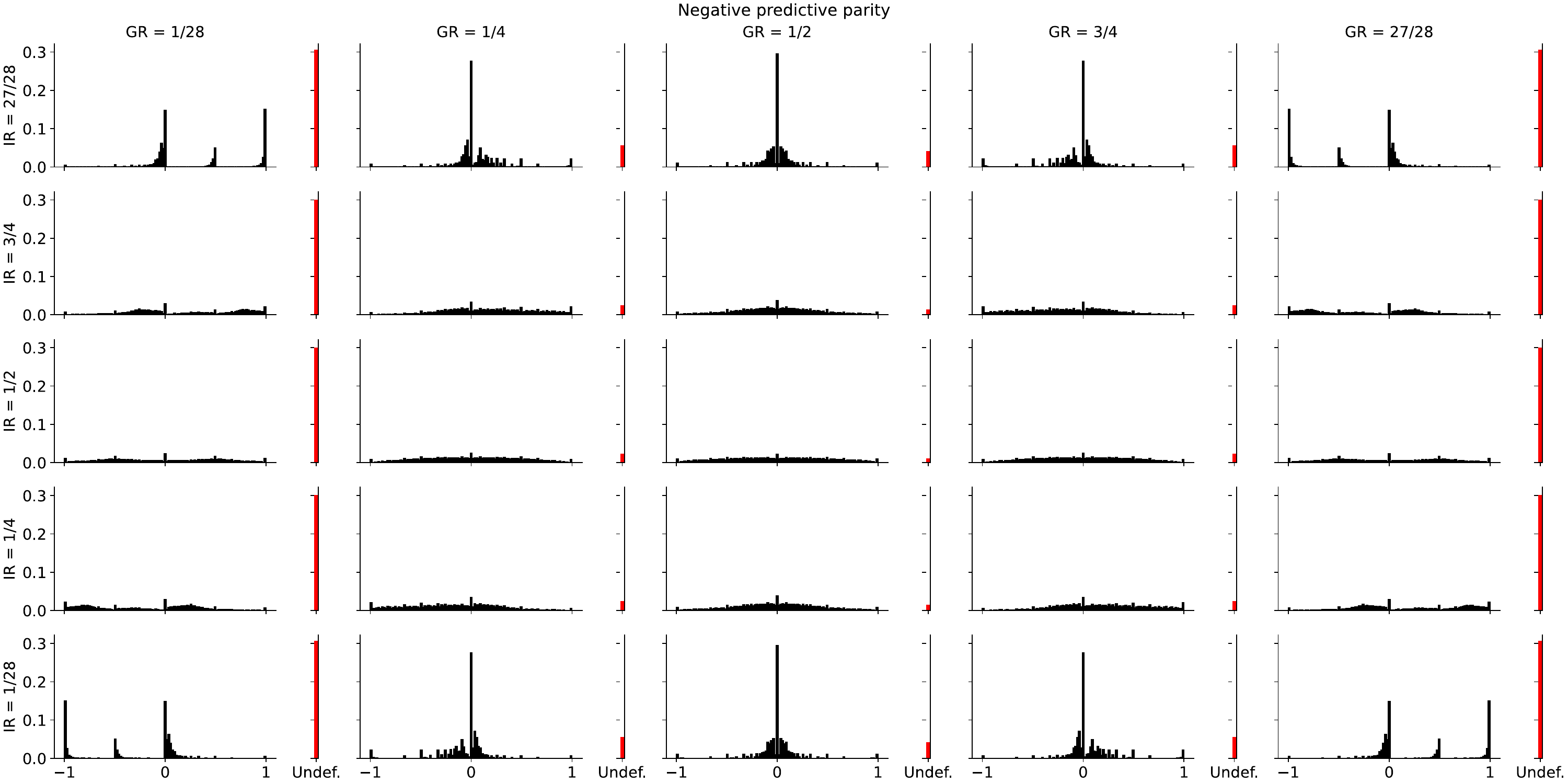}
}
\caption{Histograms of \textbf{negative predictive parity} values for selected  imbalance and group ratios. The x-axis spans between the minimum and maximum of each measure ($[-1, 1]$) and is divided into bins of equal width. The y-axis shows the probability of obtaining a given measure value. For a given measure, different panels represent varying imbalance ratios $\mathit{IR}$ (top-bottom) and group ratios $\mathit{GR}$ (left-right). The probability of undefined values (Undef.) is represented as red bars next to each histogram.}\label{fig:hist_neg_pred_parity}
\end{figure*}

\begin{figure*}[h!]
\centerline{
    \includegraphics[width=0.65\textwidth]{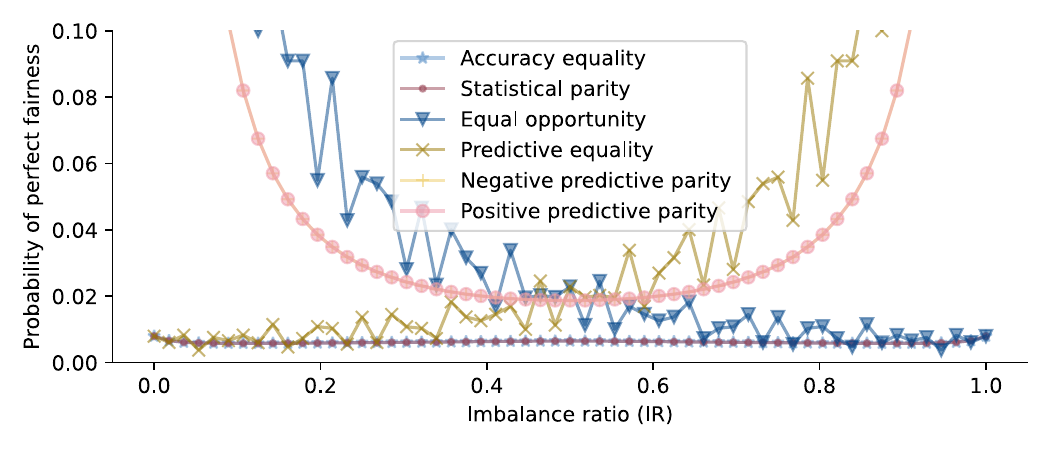}
}
\caption{Probability of perfect fairness as a function of imbalance ratio.}\label{fig:ppf_ir}
\end{figure*}

\begin{figure*}[h!]
\centerline{
    \includegraphics[width=0.65\textwidth]{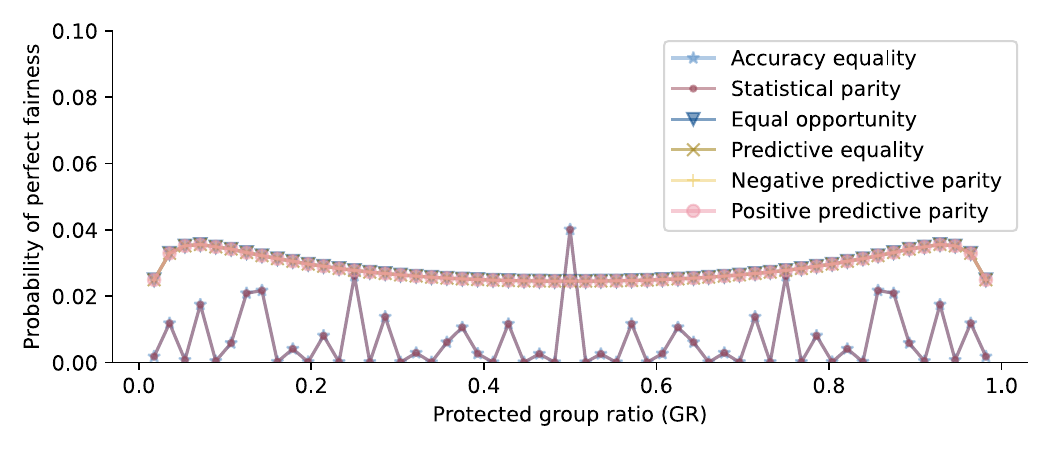}
}
\caption{Probability of perfect fairness as a function of protected group ratio.}\label{fig:ppf_gr}
\end{figure*}

\begin{figure*}[h!]
\centerline{
    \includegraphics[width=0.65\textwidth]{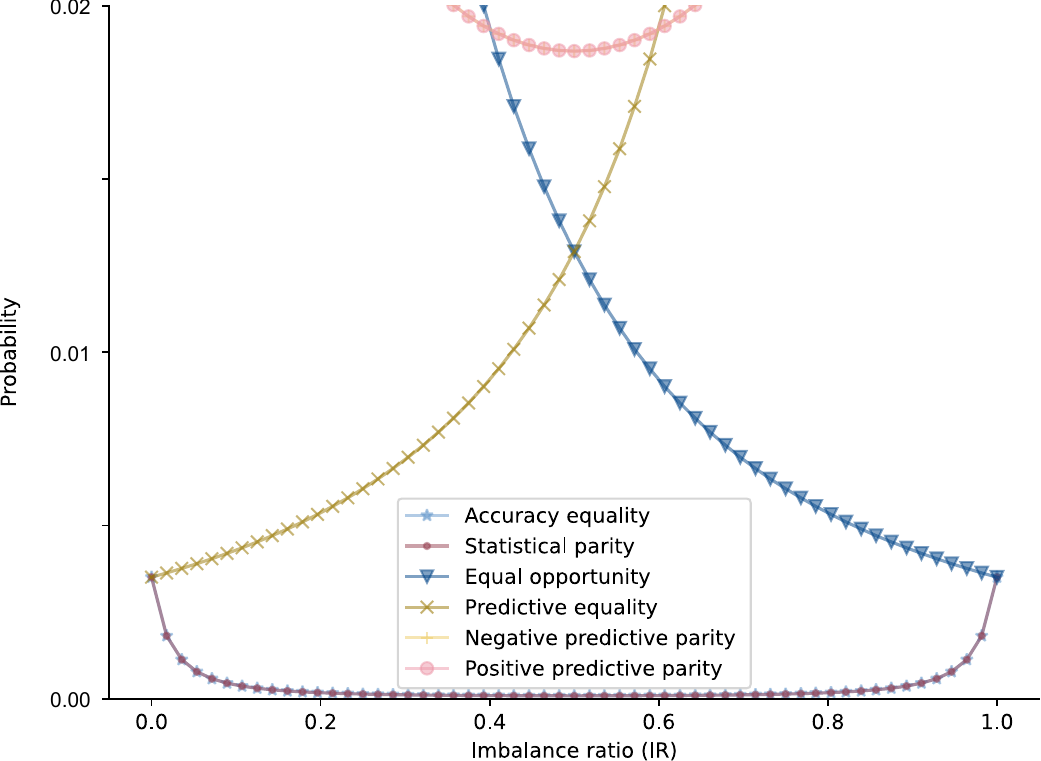}
}
\caption{Probability of undefined metric value as a function of imbalance ratio}\label{fig:nan_ir}
\end{figure*}

\begin{figure*}[h!]
\centerline{
    \includegraphics[width=0.65\textwidth]{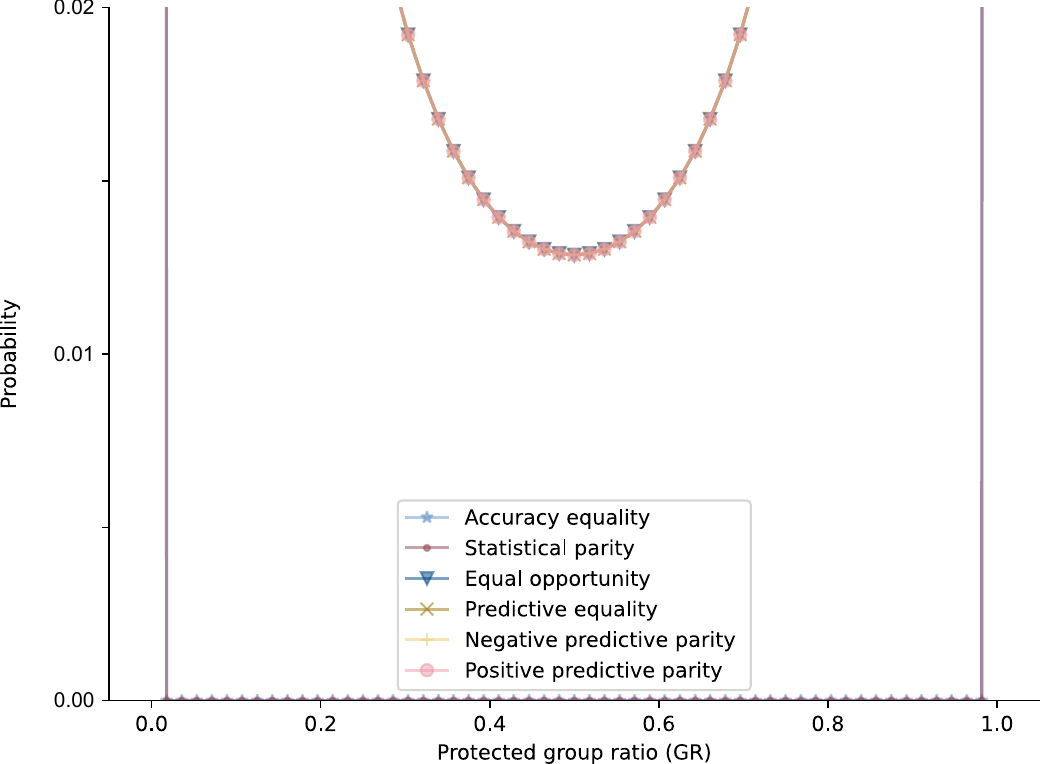}
}
\caption{Probability of undefined metric value as a function of protected group ratio}\label{fig:nan_gr}
\end{figure*}

\begin{figure*}[h!]
\centerline{
    \includegraphics[width=0.85\textwidth]{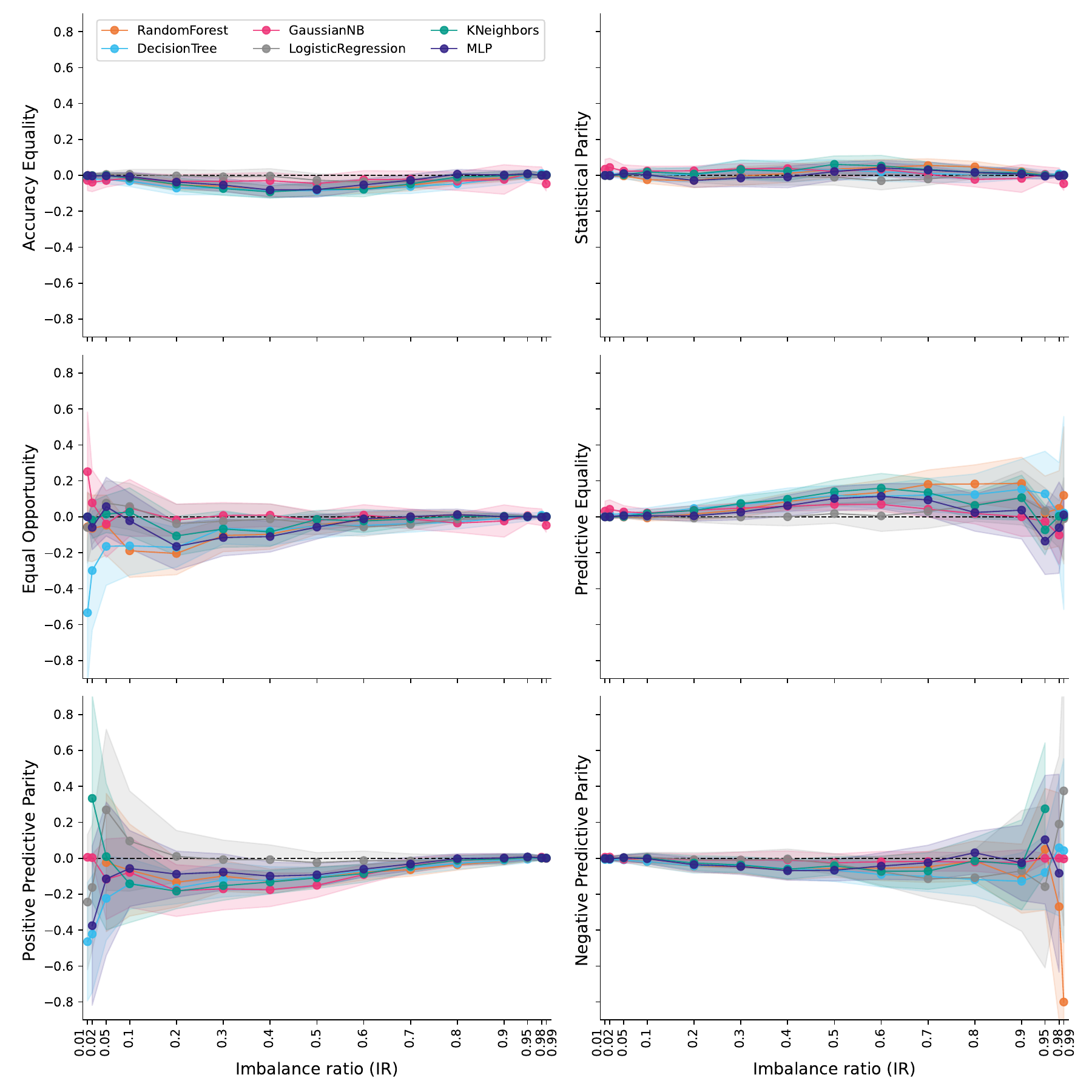}
}
\caption{Case study results: fairness achieved by the analyzed classifiers in relation to imbalance ratio.}\label{fig:cs_line_ir}
\end{figure*}

\begin{figure*}[h!]
\centerline{
    \includegraphics[width=0.85\textwidth]{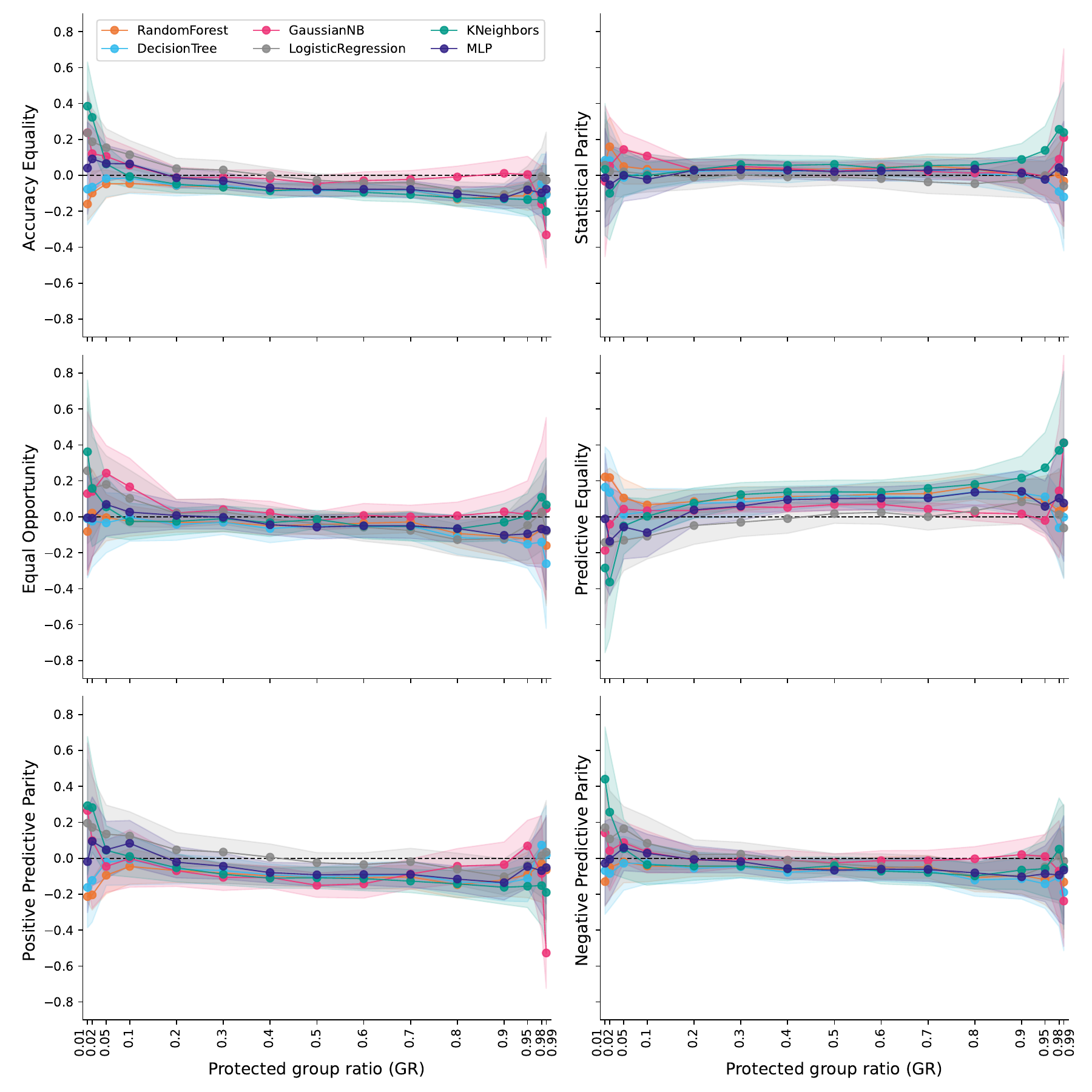}
}
\caption{Case study results: fairness achieved by the analyzed classifiers in relation to protected group ratio.}\label{fig:cs_line_gr}
\end{figure*}

\clearpage

\begin{table*}[h!]
\begin{center}
{\caption{Case study: ROC AUC scores of the considered classifiers (average and standard deviation).}\label{tab:classification_roc_auc}}
\begin{tabular}{l l c c c c c }
\toprule
$\mathit{IR}$ & $\mathit{GR}$ & RandomForest & DecisionTree & GaussianNB & LogisticRegression & KNeighbors \\
\midrule
0.01 & 0.50 & 0.514 (0.048) & 0.645 (0.136) & 0.584 (0.067) & 0.519 (0.078) & 0.500 (0.000) \\
0.02 & 0.50 & 0.511 (0.029) & 0.610 (0.065) & 0.614 (0.090) & 0.516 (0.038) & 0.508 (0.026) \\
0.05 & 0.50 & 0.588 (0.043) & 0.648 (0.048) & 0.608 (0.050) & 0.540 (0.030) & 0.534 (0.039) \\
0.10 & 0.50 & 0.640 (0.037) & 0.668 (0.035) & 0.620 (0.030) & 0.587 (0.024) & 0.591 (0.031) \\
0.20 & 0.50 & 0.725 (0.034) & 0.701 (0.036) & 0.622 (0.024) & 0.658 (0.028) & 0.668 (0.023) \\
0.30 & 0.50 & 0.795 (0.021) & 0.732 (0.024) & 0.620 (0.020) & 0.706 (0.025) & 0.733 (0.022) \\
0.40 & 0.50 & 0.817 (0.017) & 0.757 (0.019) & 0.620 (0.021) & 0.735 (0.020) & 0.760 (0.016) \\
0.50 & 0.50 & 0.830 (0.015) & 0.770 (0.021) & 0.636 (0.021) & 0.755 (0.017) & 0.781 (0.018) \\
0.60 & 0.50 & 0.822 (0.016) & 0.759 (0.026) & 0.649 (0.023) & 0.742 (0.023) & 0.773 (0.018) \\
0.70 & 0.50 & 0.804 (0.018) & 0.746 (0.026) & 0.662 (0.026) & 0.715 (0.021) & 0.749 (0.020) \\
0.80 & 0.50 & 0.767 (0.030) & 0.728 (0.031) & 0.668 (0.026) & 0.662 (0.023) & 0.713 (0.026) \\
0.90 & 0.50 & 0.683 (0.039) & 0.675 (0.041) & 0.702 (0.056) & 0.600 (0.028) & 0.635 (0.032) \\
0.95 & 0.50 & 0.624 (0.054) & 0.662 (0.050) & 0.639 (0.024) & 0.541 (0.029) & 0.569 (0.039) \\
0.98 & 0.50 & 0.553 (0.051) & 0.611 (0.073) & 0.624 (0.045) & 0.513 (0.027) & 0.501 (0.010) \\
0.99 & 0.50 & 0.568 (0.089) & 0.646 (0.117) & 0.681 (0.071) & 0.509 (0.037) & 0.500 (0.000) \\
0.50 & 0.01 & 0.792 (0.019) & 0.731 (0.022) & 0.698 (0.048) & 0.781 (0.017) & 0.746 (0.020) \\
0.50 & 0.02 & 0.789 (0.020) & 0.733 (0.023) & 0.692 (0.048) & 0.771 (0.018) & 0.736 (0.021) \\
0.50 & 0.05 & 0.795 (0.016) & 0.733 (0.023) & 0.696 (0.042) & 0.776 (0.018) & 0.745 (0.022) \\
0.50 & 0.10 & 0.795 (0.016) & 0.734 (0.021) & 0.670 (0.040) & 0.767 (0.014) & 0.744 (0.021) \\
0.50 & 0.20 & 0.802 (0.015) & 0.734 (0.017) & 0.627 (0.017) & 0.755 (0.018) & 0.743 (0.021) \\
0.50 & 0.30 & 0.807 (0.015) & 0.737 (0.022) & 0.619 (0.018) & 0.743 (0.019) & 0.752 (0.017) \\
0.50 & 0.40 & 0.819 (0.016) & 0.755 (0.020) & 0.628 (0.017) & 0.738 (0.019) & 0.766 (0.021) \\
0.50 & 0.50 & 0.830 (0.015) & 0.770 (0.021) & 0.636 (0.021) & 0.755 (0.017) & 0.781 (0.018) \\
0.50 & 0.60 & 0.829 (0.014) & 0.769 (0.020) & 0.635 (0.021) & 0.751 (0.019) & 0.777 (0.017) \\
0.50 & 0.70 & 0.835 (0.014) & 0.771 (0.018) & 0.638 (0.016) & 0.759 (0.022) & 0.788 (0.016) \\
0.50 & 0.80 & 0.830 (0.014) & 0.762 (0.018) & 0.646 (0.017) & 0.761 (0.019) & 0.787 (0.019) \\
0.50 & 0.90 & 0.845 (0.013) & 0.779 (0.022) & 0.648 (0.015) & 0.769 (0.020) & 0.799 (0.017) \\
0.50 & 0.95 & 0.851 (0.017) & 0.795 (0.020) & 0.649 (0.017) & 0.778 (0.021) & 0.805 (0.021) \\
0.50 & 0.98 & 0.855 (0.015) & 0.796 (0.019) & 0.655 (0.019) & 0.790 (0.018) & 0.811 (0.017) \\
0.50 & 0.99 & 0.850 (0.017) & 0.796 (0.020) & 0.653 (0.024) & 0.782 (0.019) & 0.806 (0.017) \\ \bottomrule
\end{tabular}
\end{center}
\end{table*}

\begin{table*}[h!]
\begin{center}
{\caption{Case study: G-mean scores of the considered classifiers (average and standard deviation).}\label{tab:classification_g_mean}}
\begin{tabular}{l l | c c c c c }
\toprule
$\mathit{IR}$ & $\mathit{GR}$  & RandomForest & DecisionTree & GaussianNB & LogisticRegression & KNeighbors \\
\midrule
0.01 & 0.50 & 0.053 (0.163) & 0.443 (0.321) & 0.503 (0.045) & 0.060 (0.194) & - \\
0.02 & 0.50 & 0.055 (0.140) & 0.452 (0.170) & 0.532 (0.104) & 0.085 (0.174) & 0.040 (0.123) \\
0.05 & 0.50 & 0.410 (0.113) & 0.562 (0.089) & 0.486 (0.111) & 0.263 (0.137) & 0.196 (0.183) \\
0.10 & 0.50 & 0.538 (0.067) & 0.615 (0.054) & 0.515 (0.055) & 0.428 (0.060) & 0.446 (0.069) \\
0.20 & 0.50 & 0.691 (0.046) & 0.679 (0.045) & 0.528 (0.044) & 0.587 (0.048) & 0.612 (0.036) \\
0.30 & 0.50 & 0.787 (0.024) & 0.723 (0.027) & 0.530 (0.031) & 0.678 (0.032) & 0.719 (0.026) \\
0.40 & 0.50 & 0.815 (0.018) & 0.755 (0.020) & 0.529 (0.032) & 0.726 (0.022) & 0.758 (0.017) \\
0.50 & 0.50 & 0.830 (0.015) & 0.769 (0.021) & 0.561 (0.029) & 0.754 (0.017) & 0.781 (0.019) \\
0.60 & 0.50 & 0.820 (0.017) & 0.757 (0.027) & 0.587 (0.030) & 0.736 (0.025) & 0.769 (0.020) \\
0.70 & 0.50 & 0.796 (0.021) & 0.739 (0.029) & 0.608 (0.031) & 0.686 (0.027) & 0.730 (0.025) \\
0.80 & 0.50 & 0.743 (0.039) & 0.711 (0.038) & 0.620 (0.030) & 0.587 (0.038) & 0.668 (0.038) \\
0.90 & 0.50 & 0.609 (0.064) & 0.625 (0.060) & 0.680 (0.067) & 0.453 (0.061) & 0.525 (0.059) \\
0.95 & 0.50 & 0.490 (0.105) & 0.588 (0.080) & 0.555 (0.021) & 0.257 (0.138) & 0.353 (0.127) \\
0.98 & 0.50 & 0.251 (0.209) & 0.450 (0.187) & 0.547 (0.030) & 0.079 (0.151) & 0.008 (0.053) \\
0.99 & 0.50 & 0.240 (0.284) & 0.462 (0.291) & 0.618 (0.058) & 0.033 (0.132) & - \\
0.50 & 0.01 & 0.791 (0.020) & 0.730 (0.023) & 0.662 (0.071) & 0.780 (0.017) & 0.745 (0.020) \\
0.50 & 0.02 & 0.788 (0.020) & 0.732 (0.023) & 0.652 (0.069) & 0.770 (0.018) & 0.734 (0.022) \\
0.50 & 0.05 & 0.795 (0.016) & 0.733 (0.024) & 0.655 (0.058) & 0.775 (0.018) & 0.744 (0.022) \\
0.50 & 0.10 & 0.795 (0.016) & 0.734 (0.022) & 0.614 (0.060) & 0.766 (0.014) & 0.742 (0.021) \\
0.50 & 0.20 & 0.801 (0.015) & 0.733 (0.017) & 0.547 (0.026) & 0.754 (0.019) & 0.741 (0.021) \\
0.50 & 0.30 & 0.806 (0.015) & 0.736 (0.022) & 0.537 (0.028) & 0.742 (0.019) & 0.751 (0.017) \\
0.50 & 0.40 & 0.819 (0.016) & 0.755 (0.020) & 0.548 (0.024) & 0.738 (0.019) & 0.765 (0.021) \\
0.50 & 0.50 & 0.830 (0.015) & 0.769 (0.021) & 0.561 (0.029) & 0.754 (0.017) & 0.781 (0.019) \\
0.50 & 0.60 & 0.829 (0.014) & 0.768 (0.020) & 0.558 (0.029) & 0.750 (0.019) & 0.776 (0.017) \\
0.50 & 0.70 & 0.835 (0.014) & 0.771 (0.019) & 0.564 (0.025) & 0.758 (0.022) & 0.788 (0.017) \\
0.50 & 0.80 & 0.830 (0.014) & 0.761 (0.018) & 0.571 (0.027) & 0.760 (0.019) & 0.787 (0.019) \\
0.50 & 0.90 & 0.844 (0.013) & 0.778 (0.022) & 0.574 (0.022) & 0.769 (0.020) & 0.799 (0.017) \\
0.50 & 0.95 & 0.850 (0.017) & 0.795 (0.020) & 0.576 (0.026) & 0.777 (0.021) & 0.805 (0.021) \\
0.50 & 0.98 & 0.855 (0.015) & 0.796 (0.019) & 0.590 (0.026) & 0.789 (0.019) & 0.811 (0.017) \\
0.50 & 0.99 & 0.850 (0.017) & 0.795 (0.020) & 0.585 (0.034) & 0.782 (0.019) & 0.806 (0.017) \\ \bottomrule
\end{tabular}
\end{center}
\end{table*}

\begin{table*}[h!]
\begin{center}
{\caption{Case study: recall of the considered classifiers (average and standard deviation).}\label{tab:classification_recall}}
\begin{tabular}{l l | c c c c c }
\toprule
$\mathit{IR}$ & $\mathit{GR}$ & RandomForest & DecisionTree & GaussianNB & LogisticRegression & KNeighbors \\
\midrule
0.01 & 0.50 & 0.029 (0.096) & 0.301 (0.273) & 0.858 (0.185) & 0.041 (0.156) & - \\
0.02 & 0.50 & 0.022 (0.058) & 0.237 (0.129) & 0.908 (0.116) & 0.037 (0.078) & 0.017 (0.053) \\
0.05 & 0.50 & 0.181 (0.087) & 0.339 (0.104) & 0.333 (0.214) & 0.088 (0.062) & 0.072 (0.079) \\
0.10 & 0.50 & 0.300 (0.075) & 0.414 (0.074) & 0.279 (0.058) & 0.190 (0.049) & 0.209 (0.064) \\
0.20 & 0.50 & 0.511 (0.069) & 0.533 (0.071) & 0.296 (0.050) & 0.366 (0.063) & 0.404 (0.050) \\
0.30 & 0.50 & 0.686 (0.047) & 0.625 (0.049) & 0.299 (0.034) & 0.514 (0.050) & 0.591 (0.042) \\
0.40 & 0.50 & 0.766 (0.035) & 0.713 (0.039) & 0.298 (0.034) & 0.624 (0.038) & 0.703 (0.033) \\
0.50 & 0.50 & 0.831 (0.028) & 0.770 (0.031) & 0.337 (0.031) & 0.743 (0.033) & 0.793 (0.030) \\
0.60 & 0.50 & 0.876 (0.022) & 0.811 (0.035) & 0.374 (0.035) & 0.832 (0.028) & 0.847 (0.029) \\
0.70 & 0.50 & 0.920 (0.018) & 0.842 (0.026) & 0.401 (0.035) & 0.914 (0.018) & 0.911 (0.022) \\
0.80 & 0.50 & 0.952 (0.015) & 0.880 (0.025) & 0.421 (0.035) & 0.967 (0.013) & 0.956 (0.012) \\
0.90 & 0.50 & 0.984 (0.007) & 0.921 (0.019) & 0.544 (0.105) & 0.989 (0.006) & 0.987 (0.007) \\
0.95 & 0.50 & 0.995 (0.004) & 0.957 (0.013) & 0.324 (0.027) & 0.996 (0.004) & 0.998 (0.002) \\
0.98 & 0.50 & 0.999 (0.002) & 0.979 (0.008) & 0.328 (0.028) & 0.997 (0.003) & 1.000 (0.001) \\
0.99 & 0.50 & 0.999 (0.001) & 0.992 (0.006) & 0.401 (0.024) & 0.999 (0.002) & 1.000 (0.000) \\
0.50 & 0.01 & 0.803 (0.026) & 0.719 (0.038) & 0.490 (0.104) & 0.798 (0.035) & 0.780 (0.033) \\
0.50 & 0.02 & 0.799 (0.038) & 0.729 (0.038) & 0.471 (0.099) & 0.781 (0.036) & 0.771 (0.037) \\
0.50 & 0.05 & 0.800 (0.026) & 0.725 (0.040) & 0.467 (0.080) & 0.785 (0.032) & 0.770 (0.036) \\
0.50 & 0.10 & 0.808 (0.033) & 0.735 (0.038) & 0.410 (0.080) & 0.765 (0.037) & 0.777 (0.036) \\
0.50 & 0.20 & 0.821 (0.031) & 0.731 (0.038) & 0.321 (0.029) & 0.746 (0.041) & 0.786 (0.032) \\
0.50 & 0.30 & 0.825 (0.026) & 0.737 (0.036) & 0.313 (0.031) & 0.727 (0.035) & 0.784 (0.031) \\
0.50 & 0.40 & 0.827 (0.032) & 0.761 (0.032) & 0.322 (0.027) & 0.725 (0.032) & 0.779 (0.039) \\
0.50 & 0.50 & 0.831 (0.028) & 0.770 (0.031) & 0.337 (0.031) & 0.743 (0.033) & 0.793 (0.030) \\
0.50 & 0.60 & 0.836 (0.024) & 0.772 (0.034) & 0.333 (0.032) & 0.739 (0.043) & 0.785 (0.026) \\
0.50 & 0.70 & 0.846 (0.029) & 0.776 (0.033) & 0.339 (0.030) & 0.743 (0.040) & 0.794 (0.025) \\
0.50 & 0.80 & 0.835 (0.034) & 0.762 (0.031) & 0.346 (0.033) & 0.754 (0.037) & 0.803 (0.031) \\
0.50 & 0.90 & 0.840 (0.025) & 0.775 (0.035) & 0.349 (0.028) & 0.773 (0.032) & 0.813 (0.028) \\
0.50 & 0.95 & 0.847 (0.026) & 0.790 (0.032) & 0.351 (0.034) & 0.784 (0.033) & 0.819 (0.035) \\
0.50 & 0.98 & 0.854 (0.026) & 0.794 (0.034) & 0.371 (0.031) & 0.803 (0.028) & 0.828 (0.031) \\
0.50 & 0.99 & 0.842 (0.024) & 0.789 (0.032) & 0.364 (0.040) & 0.788 (0.037) & 0.814 (0.030) \\ \bottomrule
\end{tabular}
\end{center}
\end{table*}

\begin{table*}[h!]
\begin{center}
{\caption{Case study: F1 scores of the considered classifiers (average and standard deviation).}\label{tab:classification_f1}}
\begin{tabular}{l l | c c c c c }
\toprule
$\mathit{IR}$ & $\mathit{GR}$ & RandomForest & DecisionTree & GaussianNB & LogisticRegression & KNeighbors \\
\midrule
0.01 & 0.50 & 0.042 (0.133) & 0.237 (0.176) & 0.025 (0.011) & 0.032 (0.098) & - \\
0.02 & 0.50 & 0.038 (0.099) & 0.220 (0.118) & 0.051 (0.016) & 0.048 (0.100) & 0.025 (0.079) \\
0.05 & 0.50 & 0.279 (0.113) & 0.308 (0.077) & 0.256 (0.108) & 0.137 (0.090) & 0.115 (0.121) \\
0.10 & 0.50 & 0.400 (0.078) & 0.386 (0.054) & 0.340 (0.068) & 0.279 (0.062) & 0.282 (0.073) \\
0.20 & 0.50 & 0.580 (0.053) & 0.517 (0.062) & 0.391 (0.051) & 0.465 (0.053) & 0.479 (0.038) \\
0.30 & 0.50 & 0.717 (0.029) & 0.625 (0.037) & 0.416 (0.041) & 0.585 (0.040) & 0.628 (0.033) \\
0.40 & 0.50 & 0.781 (0.023) & 0.710 (0.025) & 0.429 (0.041) & 0.673 (0.028) & 0.712 (0.022) \\
0.50 & 0.50 & 0.830 (0.016) & 0.770 (0.021) & 0.480 (0.036) & 0.752 (0.019) & 0.783 (0.021) \\
0.60 & 0.50 & 0.863 (0.014) & 0.808 (0.023) & 0.524 (0.038) & 0.806 (0.018) & 0.827 (0.017) \\
0.70 & 0.50 & 0.896 (0.011) & 0.845 (0.017) & 0.558 (0.037) & 0.861 (0.011) & 0.872 (0.011) \\
0.80 & 0.50 & 0.925 (0.010) & 0.885 (0.015) & 0.583 (0.035) & 0.908 (0.010) & 0.914 (0.010) \\
0.90 & 0.50 & 0.958 (0.008) & 0.928 (0.011) & 0.691 (0.088) & 0.951 (0.008) & 0.954 (0.009) \\
0.95 & 0.50 & 0.977 (0.006) & 0.961 (0.007) & 0.488 (0.030) & 0.973 (0.006) & 0.976 (0.006) \\
0.98 & 0.50 & 0.990 (0.003) & 0.981 (0.004) & 0.493 (0.031) & 0.988 (0.003) & 0.989 (0.003) \\
0.99 & 0.50 & 0.995 (0.003) & 0.992 (0.003) & 0.571 (0.025) & 0.994 (0.002) & 0.994 (0.002) \\
0.50 & 0.01 & 0.795 (0.020) & 0.728 (0.027) & 0.613 (0.092) & 0.785 (0.017) & 0.755 (0.021) \\
0.50 & 0.02 & 0.791 (0.022) & 0.732 (0.026) & 0.599 (0.089) & 0.773 (0.020) & 0.745 (0.023) \\
0.50 & 0.05 & 0.797 (0.019) & 0.732 (0.028) & 0.603 (0.074) & 0.779 (0.020) & 0.752 (0.023) \\
0.50 & 0.10 & 0.798 (0.018) & 0.734 (0.025) & 0.550 (0.077) & 0.766 (0.017) & 0.752 (0.021) \\
0.50 & 0.20 & 0.805 (0.017) & 0.733 (0.021) & 0.462 (0.034) & 0.752 (0.020) & 0.753 (0.021) \\
0.50 & 0.30 & 0.809 (0.016) & 0.735 (0.026) & 0.450 (0.036) & 0.737 (0.021) & 0.758 (0.018) \\
0.50 & 0.40 & 0.820 (0.018) & 0.756 (0.023) & 0.463 (0.030) & 0.734 (0.019) & 0.768 (0.023) \\
0.50 & 0.50 & 0.830 (0.016) & 0.770 (0.021) & 0.480 (0.036) & 0.752 (0.019) & 0.783 (0.021) \\
0.50 & 0.60 & 0.831 (0.015) & 0.770 (0.024) & 0.477 (0.038) & 0.749 (0.024) & 0.780 (0.018) \\
0.50 & 0.70 & 0.837 (0.015) & 0.772 (0.020) & 0.484 (0.032) & 0.754 (0.025) & 0.789 (0.018) \\
0.50 & 0.80 & 0.831 (0.017) & 0.762 (0.021) & 0.493 (0.035) & 0.759 (0.023) & 0.791 (0.022) \\
0.50 & 0.90 & 0.843 (0.014) & 0.777 (0.025) & 0.497 (0.029) & 0.769 (0.018) & 0.801 (0.017) \\
0.50 & 0.95 & 0.850 (0.019) & 0.795 (0.023) & 0.500 (0.034) & 0.780 (0.021) & 0.808 (0.023) \\
0.50 & 0.98 & 0.855 (0.016) & 0.796 (0.022) & 0.518 (0.033) & 0.793 (0.018) & 0.815 (0.017) \\
0.50 & 0.99 & 0.849 (0.018) & 0.794 (0.023) & 0.511 (0.043) & 0.783 (0.020) & 0.808 (0.021) \\ \bottomrule
\end{tabular}
\end{center}
\end{table*}

\end{document}